\documentclass[preprint,12pt]{elsarticle}
\biboptions{numbers,sort&compress} 
\usepackage{amsmath,amssymb,amsfonts}
\usepackage{graphicx}
\usepackage{textcomp}
\usepackage{xcolor}
\usepackage{booktabs}
\usepackage{multirow}
\usepackage{makecell}
\usepackage{hyperref}
\usepackage{caption}
\usepackage{float}
\usepackage[nameinlink]{cleveref}
\crefname{figure}{Fig.}{Figs.}
\Crefname{figure}{Fig.}{Figs.}
\Crefname{table}{Table}{Tables}
\usepackage{tabularx}      
\usepackage{array}        
\usepackage{makecell}      
\usepackage{pifont}        
\usepackage{booktabs}

\usepackage{algorithm}
\usepackage{algpseudocode} 

\makeatletter
\renewcommand{\ALG@name}{Algorithm} 
\algrenewcommand\algorithmicrequire{\textbf{Input:}}
\algrenewcommand\algorithmicensure{\textbf{Output:}} 
\makeatother

\hypersetup{
    colorlinks=true,
    linkcolor=blue,
    citecolor=blue,
    urlcolor=cyan,
}

\journal{ }

\begin{document}

\begin{frontmatter}

\title{Fusing Urban Structure and Semantics: A Conditional Diffusion Model for Cross-City OD Matrix Generation}

\author[aff1]{Bin Chen\fnref{equal}}
\author[aff1]{Zhuoya Meng\fnref{equal}}
\author[aff2]{Fang Yang}
\author[aff2]{Runkang Guo}
\author[aff3]{Jingtao Ding}
\author[aff4]{Yin Zhang}
\author[aff2]{Chuan Ai\corref{cor1}}
\author[aff2]{Zhengqiu Zhu\corref{cor1}}

\cortext[cor1]{Corresponding author, zhuzhengqiu12@nudt.edu.cn}
\fntext[equal]{Equal Contribution}

\address[aff1]{Institute of Intelligent Computing, University of Electronic Science and Technology of China (UESTC), Chengdu, 611731, Sichuan, China}
\address[aff2]{College of Systems Engineering, National University of Defense Technology, Changsha, 410073, Hunan, China}
\address[aff3]{Department of Electronic Engineering, Beijing National Research Center for Information Science and Technology (BNRist), Tsinghua University, Beijing, China}
\address[aff4]{School of Information and Communication Engineering, University of Electronic Science and Technology of China (UESTC), Chengdu, 611731, Sichuan, China}

\begin{abstract}
Accurate modeling of commuting flows is important for urban governance, traffic planning, and resource allocation. However, the combined influence of individual intentions, geographic constraints, and social dynamics leads to considerable heterogeneity in commuting patterns, making it difficult to develop generation models that generalize across cities. To address this issue, we propose SEDAN, a \textbf{S}tructure-\textbf{E}nhanced \textbf{D}iffusion model conditioned on \textbf{A}ttributed \textbf{N}odes for generalizable OD matrix generation. SEDAN models a city as an attributed graph. Each region is treated as a node with demographic and point-of-interest features, and commuting flows are modeled as weighted edges. Adjacency and distance matrices are incorporated to characterize spatial structure. Based on this representation, we design a fusion mechanism within SEDAN to jointly model semantic information and spatial information. Regional semantic attributes are used to model latent travel demand through graph-transformer-based node interactions, while spatial structure is injected into the generation process as explicit constraints. The adjacency matrix guides attention weights to strengthen interactions between neighboring regions. Meanwhile, the distance matrix serves as a diffusion condition to capture spatial proximity and travel impedance. The fusion of urban semantics and spatial constraints enables SEDAN to generate OD matrices that are both behaviorally plausible and geographically coherent. Experiments on real-world OD datasets from U.S. cities show that SEDAN achieves a 7.38\% improvement in RMSE over the state-of-the-art baseline, WEDAN. It also remains robust across heterogeneous urban scenarios and varying structural patterns. Our work provides an effective and generalizable solution for commuting OD matrix generation. The code is available at https://anonymous.4open.science/r/SEDAN.
\end{abstract}


\end{frontmatter}

\section{Introduction}
Commuting behavior, as a core component of daily travel, reflects the spatiotemporal relationship between where people live and where they work. The origin–destination (OD) matrix represents commuting flows between urban regions. It can be modeled as a directed, weighted graph, where nodes denote regions and edges denote the intensity of OD flows~\cite{saberi2017complex, saberi2018complex,rong2024interdisciplinary}. As urban systems grow increasingly complex, transportation and land-use planning face escalating challenges due to longer commutes, job–housing imbalances, and intensified peak-hour congestion. OD flows serve as a fundamental representation of commuting behavior and provide important insights for urban zoning and transport planning~\cite{zeng2024estimating,credit2023method,yang2020understanding}. Therefore, accurate and generalizable OD flow modeling and generation have emerged as critical tasks in urban computing.

Existing OD modeling methods can be broadly categorized into mechanism-driven and data-driven approaches. Traditional mechanism-driven models, such as gravity models~\cite{zipf1946p} and radiation models~\cite{simini2012universal}, primarily describe human mobility patterns as functions of distance and population. Later variants, such as the attraction-constrained gravity model~\cite{griffith2016constrained}, the doubly-constrained gravity model~\cite{cai2021doubly}, and the universal opportunity model~\cite{liu2020universal}, better capture diverse mobility contexts and urban structural characteristics. However, they still struggle to capture complex spatial heterogeneity and nonlinear dependencies, which limits their performance across diverse urban contexts.

With recent advances in artificial intelligence, data-driven learning has become the dominant paradigm for modeling urban OD flows. Early studies employed supervised architectures, such as Convolutional Neural Networks (CNNs)~\cite{rong2019deep}, Recurrent Neural Networks (RNNs)~\cite{rong2024learning}, and Graph Neural Networks (GNNs)~\cite{rong2021inferring}. These methods perform well when data are abundant. However, many cities suffer from a lack of mobility data due to limited sensing infrastructure, motivating the exploration of generative approaches. Models such as VAEs~\cite{huang2019variational}, GANs~\cite{ouyang2023citytrans}, and diffusion models~\cite{rong2023complexity} can learn underlying mobility patterns from data-rich cities and generate OD flows for regions with scarce historical data. To improve cross-city generalization, transfer learning methods~\cite{he2020human,rong2023goddag} align feature spaces and transfer knowledge across domains. However, their effectiveness often depends on domain similarity. Unlike macro-level OD flow generation, LLM-based approaches~\cite{bhandari2024urban} focus on modeling individual-level travel behaviors and leverage the pretrained knowledge of large language models to synthesize realistic travel survey data.

Although recent methods have shown promising results, they often struggle to simultaneously achieve high modeling accuracy and robust cross-city generalization. Urban commuting flows are shaped by multiple factors, including regional functions, urban structure, individual travel preferences, and socioeconomic conditions. Consequently, we regard the generation of OD matrices as a task driven by the coupling of multi-source heterogeneous information. Specifically, we focus on three core challenges:
\begin{enumerate}
    \item \
    Urban population mobility is influenced by a wide range of factors, including diverse regional attributes, urban topology, and spatial relationships between regions. Designing effective mechanisms to fuse this heterogeneous information for modeling population mobility patterns remains a critical challenge.

    \item \
    Urban commuting flows are influenced by multi-scale spatial dependencies, including local interactions between adjacent regions and long-range connections between residential areas and urban centers. Accurately capturing both local and global spatial structures remains a challenge in fine-grained OD flow modeling.
    
    \item \
    Different cities exhibit significant differences in population distribution, land use patterns, transportation network topology, and mobility behaviors. These discrepancies greatly hinder the transferability and applicability of models to new cities.

\end{enumerate}

Despite the complex heterogeneity across cities, commuting flows consistently exhibit fundamental spatial regularities, such as distance-decay effects and high flow concentration between adjacent regions, as detailed in ~\ref{sec:Spatial_Regularities}. Existing methods often mix city-specific semantics with universal spatial priors in an implicit representation, which can hinder the preservation of these spatial patterns. Instead, we treat spatial information as an explicit structural constraint and incorporate spatial priors into the generation process to enforce the structural consistency of OD flows. We first formulate the OD matrix as a directed, weighted graph and define its generation as a spatially constrained edge-weight prediction task. We then propose SEDAN, a \textbf{S}tructure-\textbf{E}nhanced \textbf{D}iffusion model conditioned on \textbf{A}ttributed \textbf{N}odes, to enhance structural modeling and enable cross-city generalization. This is achieved through a heterogeneous information fusion mechanism, where semantic information models latent commuting demand and spatial information provides explicit structural constraints during generation. Specifically, a graph transformer captures the nonlinear relationships among urban regions to model latent commuting demand based on demographic and functional features. Meanwhile, during the diffusion process, two complementary spatial priors are introduced as structural constraints. The adjacency matrix guides the attention mechanism to preserve local topological connectivity, while the distance matrix provides edge-level diffusion conditioning to capture global spatial proximity and distance-decay effects. To mitigate scale disparities across cities and long-tail effects, we apply a logarithmic transformation to stabilize training. The main contributions are summarized as follows:
\begin{enumerate}
    \item 

    We propose SEDAN, a structure-enhanced graph diffusion model that reformulates OD matrix generation as a semantics-driven task under spatial structural constraints. By explicitly fusing regional attributes and spatial priors, it accurately models and generates complex commuting flows.

    \item
    We introduce a spatially guided denoising mechanism for joint modeling of multi-scale spatial dependencies. It combines adjacency-based attention to capture local connectivity and distance-based diffusion conditioning to model global spatial proximity, thereby improving the geographic consistency of generated OD matrices.
    \item
    We develop a graph transformer-based denoising network guided by spatial constraints. It learns a stable mapping between spatial structures and flow distributions, enabling robust generation across cities with diverse scales and morphologies.
    \item 
    Experiments on the LargeCommuingOD dataset~\cite{ronglarge} show that SEDAN outperforms state-of-the-art baselines on key metrics, including CPC, RMSE and JSD. The model also remains robust across cities with diverse urban structures and different levels of urban heterogeneity.
\end{enumerate}

\section{Related Works}

\subsection{Data-Driven Approaches for Modeling OD Flows}

Origin-Destination (OD) flow refers to the movement of entities (e.g., people, vehicles) from an origin to a destination during a specified time period. Driven by the growing availability of large-scale urban data, OD modeling has evolved from rule-based methods to data-driven approaches, including predictive models, transfer learning and generative models.

Supervised predictive models typically leverage regional attributes and historical flow data to predict OD flows.  Machine learning methods, such as gradient-boosted regression trees~\cite{robinson2018machine} and random forests~\cite{pourebrahim2019trip}, show improved performance over conventional models. More recent approaches incorporate spatial and functional information: GMEL~\cite{liu2020learning} exploits adjacency relations to enhance representation learning, while Rong et al.~\cite{rong2021inferring} capture spatiotemporal variations in population distribution and incorporate POI data to infer OD flows. However, most predictive models are trained on cities with available OD data and applied to predict flows in target cities, which often results in cross-city errors~\cite{simini2021deep,zeng2022causal}.

To mitigate errors in cross-city OD generation, transfer learning has been explored to improve model transferability by projecting data from different cities into a shared latent space. For instance, GODDAG~\cite{rong2023goddag} combines graph neural networks with domain adversarial learning to transfer mobility patterns from a source city to a target city for OD flow generation. However, their transfer performance often declines when substantial structural differences exist between source and target cities.  

Compared with the above methods, generative models~\cite{kingma2013auto,goodfellow2014generative,ho2020denoising} have attracted increasing attention due to their low dependence on privacy-sensitive data and effectiveness in sparse-data scenarios. NetGAN~\cite{bojchevski2018netgan}, a representative method, generates graph structures by learning random walk sequences, effectively preserving topological features while enabling link prediction. However, it struggles to explicitly model node attributes and geographic correlations. Our approach explicitly incorporates both node attributes (i.e., demographics and POIs) and their geographic spatial correlations into the generation process, enabling more accurate and spatially-consistent modeling of OD flow distributions.  

\subsection{Diffusion Models for OD Flow Generation}
Diffusion models~\cite{ho2020denoising} have emerged as powerful generative frameworks, achieving remarkable success across various domains, including image~\cite{rombach2022high,croitoru2023diffusion}, audio~\cite{liu2022diffsinger,yang2023diffsound}, and text generation~\cite{li2022diffusion,gong2022diffuseq,wu2023ar}. Recently, their integration with graph learning has opened up new opportunities for urban OD flow generation.

Among existing studies, DiffODGen~\cite{rong2023complexity} models the spatial distribution of OD matrices from a network perspective. It employs a cascaded diffusion mechanism to separately learn graph topology and edge weights, enhancing the representation of sparse OD structures. Nonetheless, it has limited capacity to model intricate spatial interactions across urban areas. More recently, by embedding geographic attribute learning into a denoising diffusion model, WEDAN~\cite{ronglarge} achieves more accurate OD matrix generation. It has become one of the strongest baseline models in OD flow generation. In addition, GlODGen\cite{rong2025satellites} generates commuting OD flows using remote sensing imagery, avoiding reliance on multi-source urban data.

Despite these advancements, most existing models fail to incorporate spatial adjacency and distance constraints, limiting their capacity to model urban spatial interactions in different urban contexts. We integrate adjacency and geographic distance into the diffusion process, which preserves local and global spatial correlations and enhances cross-city transferability. \Cref{tab:method_comparison} compares the key properties of SEDAN with other representative methods.

\begin{table}[ht]
\centering
\scriptsize
\renewcommand{\arraystretch}{1.12}
\setlength{\tabcolsep}{3pt}
\caption{Comparison of representative methods on key properties.}
\label{tab:method_comparison}
\begin{tabularx}{\linewidth}{l l l l l c c c}
\toprule
\textbf{Method} & \textbf{Classif.} & \textbf{Rep.} &
\makecell[l]{\textbf{Information}} &
\makecell[l]{\textbf{Spatial}} &
\makecell[l]{\textbf{Temporal}} &
\textbf{Modeling} & \textbf{Gener.} \\
\toprule

GMEL~\cite{liu2020learning} &
\makecell[l]{Predictive\\Model} & Graph &
\makecell[l]{Infrastructure,\\Land Use} &
\makecell[l]{Adjacency,\\Distance} &
\textcolor{red!70!black}{\ding{55}} &
Pair-wise &
\textcolor{red!70!black}{\ding{55}} \\
\addlinespace[5pt]

GSTE-DF~\cite{rong2021inferring} &
\makecell[l]{Predictive\\Model} & Graph &
\makecell[l]{Population\\Distribution,\\POIs} &
\makecell[l]{Adjacency} &
\textcolor{green!60!black}{\ding{51}} &
Pair-wise &
\textcolor{red!70!black}{\ding{55}} \\
\addlinespace[5pt]

Deep Gravity~\cite{simini2021deep} &
\makecell[l]{Predictive\\Model} & Grid &
\makecell[l]{Population,\\Land Use,\\Road Network,\\POIs} &
\makecell[l]{Distance} &
\textcolor{red!70!black}{\ding{55}} &
Pair-wise &
\textcolor{green!60!black}{\ding{51}} \\
\addlinespace[5pt]

SIRI~\cite{zeng2022causal} &
\makecell[l]{Predictive\\Model} & Grid &
\makecell[l]{Demographics,\\POIs} &
\makecell[l]{Distance} &
\textcolor{red!70!black}{\ding{55}} &
Pair-wise &
\textcolor{red!70!black}{\ding{55}} \\
\addlinespace[5pt]

NetGAN~\cite{bojchevski2018netgan} &
GAN & Graph &
\makecell[l]{Adjacency-Based\\Random Walk} &
\makecell[l]{--} &
\textcolor{red!70!black}{\ding{55}} &
Holistic &
\textcolor{red!70!black}{\ding{55}} \\
\addlinespace[5pt]

DiffODGen~\cite{rong2023complexity} &
\makecell[l]{Diffusion\\Model} & Graph &
\makecell[l]{Demographics,\\POIs} &
\makecell[l]{Distance} &
\textcolor{red!70!black}{\ding{55}} &
Holistic &
\textcolor{green!60!black}{\ding{51}} \\
\addlinespace[5pt]

WeDAN~\cite{ronglarge} &
\makecell[l]{Diffusion\\Model} & Graph &
\makecell[l]{Demographics,\\POIs} &
\makecell[l]{Distance} &
\textcolor{red!70!black}{\ding{55}} &
Holistic &
\textcolor{green!60!black}{\ding{51}} \\
\addlinespace[5pt]

SEDAN (ours) &
\makecell[l]{Diffusion\\Model} & Graph &
\makecell[l]{Demographics,\\POIs} &
\makecell[l]{Adjacency,\\Distance} &
\textcolor{red!70!black}{\ding{55}} &
Holistic &
\textcolor{green!60!black}{\ding{51}} \\
\bottomrule
\end{tabularx}

\caption*{\footnotesize
\textit{Abbreviations:} Classif.=Classification; Rep.=Representation; Gener.=Generalization.
\\
\textit{Notes:} “Modeling” denotes the modeling granularity, where \textbf{Pair-wise} independently predicts each OD pair, while \textbf{Holistic} learns the global distribution jointly.
“Temporal” indicates whether temporal dynamics are modeled. 
“Generalization” reflects whether the model demonstrates cross-city transferability.
}
\end{table}

\section{Problem Formulation for Cross-City OD Generation}

\textbf{Urban Region:}
An urban region refers to an administrative unit (e.g., census tract) representing a specific spatial area. Suppose a city is partitioned into $N$ such regions. The set of urban regions is defined as:
\begin{equation}
    \mathcal{R} = \{r_1, r_2, \ldots, r_N\}.
    \label{eq:regions}
\end{equation}

\textbf{Regional Features:} Regional features describe the characteristics of an urban region, including demographic attributes and POIs. The feature vector for region $r_i$ is defined as:
\begin{equation}
\boldsymbol{X}_i = [\boldsymbol{X}_i^{(1)}, \boldsymbol{X}_i^{(2)}] \in \mathbb{R}^d, \quad d = d_1 + d_2,
\end{equation}
where $\boldsymbol{X}_i^{(1)} \in \mathbb{R}^{d_1}$ represents demographic features and $\boldsymbol{X}_i^{(2)} \in \mathbb{R}^{d_2}$ represents POI features. The full feature matrix for the city is:
\begin{equation}
\mathbf{X}_\mathcal{R} = [\boldsymbol{X}_1, \boldsymbol{X}_2, \ldots, \boldsymbol{X}_N]^\top \in \mathbb{R}^{N \times d}.
\end{equation}

\textbf{Commuting OD Flow:}  
Commuting OD flow represents the number of commuters moving between urban regions. It is directional, reflecting both outflow and inflow. The commuting OD matrix is defined as:
\begin{equation}
    \mathbf{F} \in \mathbb{R}^{N \times N},
    \label{eq:od_matrix}
\end{equation}
where $F_{ij}$ denotes the number of commuters traveling from urban region $r_i$ to region $r_j$.

\textbf{Cross-City OD Generation Problem:} Given a set of source cities with observed OD matrices and urban characteristics, the goal is to learn a transferable model that generates the commuting OD matrix $\mathbf{F}$ for an unseen target city based only on its regional features and urban spatial structure.

\section{Methodology}

The overall architecture of SEDAN comprises two core components: an urban  characteristics embedding module and a graph transformer. The embedding module encodes regional attributes (population demographics and POIs) and spatial information (adjacency and distance matrices). The graph transformer captures complex nonlinear relationships among urban regions, modeling how regional functionality and population characteristics jointly shape latent travel demand under spatial adjacency and distance constraints. Together, these two components enable SEDAN to generate OD matrices that are structurally consistent and behaviorally plausible.

\subsection{Directed Weighted Graph Modeling of OD Flows}

To investigate the relationship between urban spatial structures and commuting patterns, the commuting OD matrix is modeled as a directed weighted graph ~\cite{saberi2017complex}
\( G = (\mathcal{V}, \mathcal{E}, \mathcal{W}) \). The city is divided into \( n = |\mathcal{V}| \) discrete regions, where
\( \mathcal{V} = \{v_1, v_2, \ldots, v_n\} \) denotes the set of nodes, each of which represents an urban region;
\( \mathcal{E} \) is the set of directed edges representing commuting between regions; 
and \( \mathcal{W} \) denotes the weights of the edges, corresponding to the volume of commuting flows.

Each node $v_i \in \mathcal{V}$ corresponds to a region $r_i$ characterized by a feature vector $\boldsymbol{X}_i$. The complete set of regional features constitutes the matrix
$\mathbf{X}_\mathcal{R} = \{ \boldsymbol{X}_i \mid v_i \in \mathcal{V} \}$.

The spatial relationships among urban regions are characterized by a distance matrix $\mathbf{D}$ and an adjacency matrix $\mathbf{A}$. The distance matrix $\mathbf{D}=\{{{d}_{ij}}|{{r}_{i}},{{r}_{j}}\in \mathcal{R}\}$ represents the geographic distance ${{d}_{ij}}$ between regions ${{r}_{i}}$ and ${{r}_{j}}$. The adjacency matrix $\mathbf{A}=[{{a}_{ij}}]\in {{\mathbb{R}}^{|\mathcal{V}|\times |\mathcal{V}|}}$ is a binary matrix indicating whether two regions are geographically adjacent: ${{a}_{ij}}=1$ if ${{r}_{i}}$ and ${{r}_{j}}$ are adjacent, and ${{a}_{ij}}=0$ otherwise.

The urban context representation of the city, denoted as ${{\mathcal{C}}_{\mathcal{R}}}$, consists of three components: the node feature ${{\mathbf{X}}_{\mathcal{R}}}$, the distance matrix $\mathbf{D}$ and the adjacency matrix $\mathbf{A}$. Collectively, they capture demographic and POI-related attributes, the spatial configuration of the city, and the topological relationships among regions.
\begin{figure}[htbp]
    \centering
    \includegraphics[width=\linewidth]{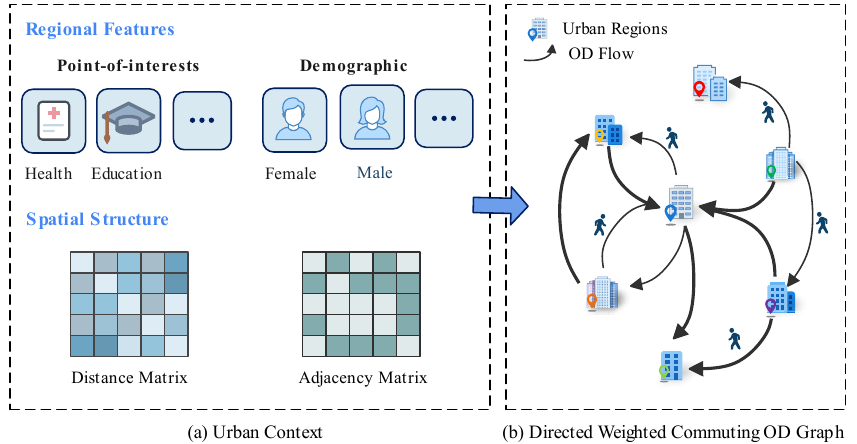} 
    \caption{Construction of the directed weighted graph for commuting OD}
    \label{fig:construction}
\end{figure}

As illustrated in~\cref{fig:construction}, the OD generation task is formulated as a conditional probabilistic modeling problem. Given the urban context representation ${{\mathcal{C}}_{\mathcal{R}}}$, the objective is to learn the distribution:
\begin{equation}
\mathcal{P}_\theta(\mathcal{E}, \{w_e \mid e \in \mathcal{E}\} \mid {{\mathcal{C}}_{\mathcal{R}}}),
\label{eq:od_prob}
\end{equation}
 which models how spatial structure and regional attributes jointly influence the commuting flows.

\subsection{Spatially-Guided Diffusion on Weighted Edges}
\begin{figure}[htbp]
    \centering
    \includegraphics[width=\linewidth]{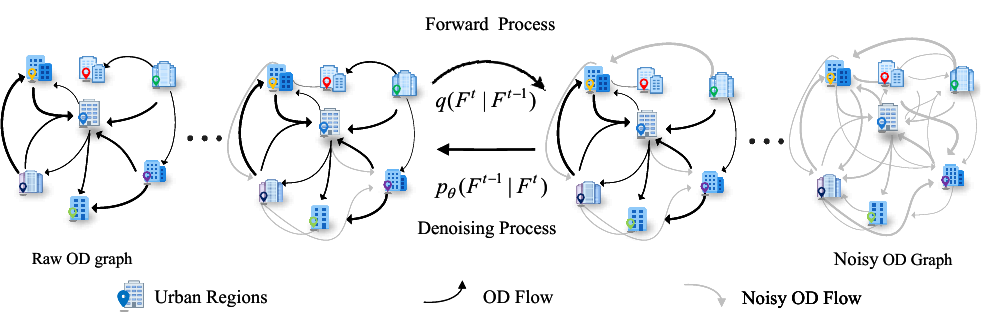} 
    \caption{Generation process of the commuting OD matrix}
    \label{fig:Generation_process}
\end{figure}

To effectively model how urban functional semantics and spatial structures jointly shape commuting flows, SEDAN employs a conditional diffusion framework that generates OD matrices under explicit spatial constraints. In real-world, flow patterns are influenced by both global distance decay and local network connectivity.
Accordingly, SEDAN conditions the diffusion process on two spatial priors: the adjacency matrix, which preserves local connectivity, and the distance matrix, which maintains global proximity.
\cref{fig:Generation_process} illustrates the two-stage process, from forward diffusion to reverse denoising.

In the forward diffusion stage, noise is incrementally added to transform the OD graph from a structured state into a fully corrupted representation. Noise is added independently to each edge, enabling precise modeling of flow values for every OD pair $(i,j)$. The process is described as:
\begin{equation}
    q(F_{ij}^{t}|F_{ij}^{t-1})=\mathcal{N}\left( F_{ij}^{t};\sqrt{1-{{\beta }_{t}}}F_{ij}^{t-1},{{\beta }_{t}}\mathbf{I} \right),
    \label{eq:forward_diffusion1}
\end{equation}
\begin{equation}
    q(F_{ij}^{1},...,F_{ij}^{T}|F_{ij}^{0})=\underset{t=1}{\overset{T}{\mathop \prod }}\,q(F_{ij}^{t}|F_{ij}^{t-1})
    \label{eq:forward_diffusion2},
\end{equation}
where $F_{ij}^{t}$ denotes the flow value of OD pair $(i,j)$ at time step $t$, and ${{\beta }_{t}}$ is the noise scale parameter that controls the intensity of noise injection. As $t$ increases, the original OD matrix is gradually perturbed by Gaussian noise, eventually converging to a standard normal distribution.

The denoising process serves as the inverse of forward diffusion, aiming to iteratively recover the true OD matrix from a noisy state. Unlike the forward process that adds noise independently to each edge, the reverse diffusion reconstructs the entire OD matrix collectively, thereby preserving the global spatial configuration and relationships among regions.

To ensure that the generated OD matrix is consistent with the spatial characteristics of the target city, SEDAN integrates a spatial conditioning mechanism. During reverse diffusion, structural information of the target city is injected into the denoising network as a conditional input, enabling the model to better capture spatial semantics and enhance the accuracy of OD flow reconstruction. The conditional probability of the reverse diffusion step is formulated as:

\begin{equation}
   p_{\theta}\!\left(\mathbf{F}^{t-1}\mid \mathbf{F}^{t},{\mathcal{C}}_{\mathcal{R}}\right)
   = \mathcal{N}\!\left(\mathbf{F}^{t-1};\, {\mu}_{\theta}(\mathbf{F}^{t},t,{\mathcal{C}}_{\mathcal{R}}),\, (1-\bar{\alpha}_{t})\,\mathbf{I}\right),
    \label{eq:Denoising_diffusion1}
\end{equation}
\begin{equation}
\mu_{\theta}(\mathbf{F}^{t},\, t,\, \mathcal{C}_{\mathcal{R}})
= \frac{1}{\sqrt{\alpha_t}}
\left(
\mathbf{F}^{t}
- \frac{\beta_t}{\sqrt{1-\bar{\alpha}_t}}\,
\epsilon_{\theta}(\mathbf{F}^{t},\, t,\, \mathcal{C}_{\mathcal{R}})
\right)
\label{eq:Denoising_diffusion2}
\end{equation}
where \(t\) is the diffusion step, \(\alpha_{t}\in(0,1)\) is the variance preserving coefficient, \(\beta_{t}=1-\alpha_{t}\), and \(\bar{\alpha}_{t}=\prod_{s=1}^{t}\alpha_{s}\). \(\mathbf{F}^{t}\in\mathbb{R}^{N\times N}\) denotes the noisy OD matrix at step \(t\), \({\epsilon}_{\theta}(\cdot)\) is the noise predicted by the denoising network with parameters \(\theta\),  $\mathcal{C}_{\mathcal{R}}$ is the urban context ( the node feature, the distance matrix  and the adjacency matrix ), and \(\mathbf{I}\) is the identity matrix.

\subsection{Structure-Enhanced Denoising Network}
\begin{figure}[ht]
\centering
\includegraphics[width=\linewidth]{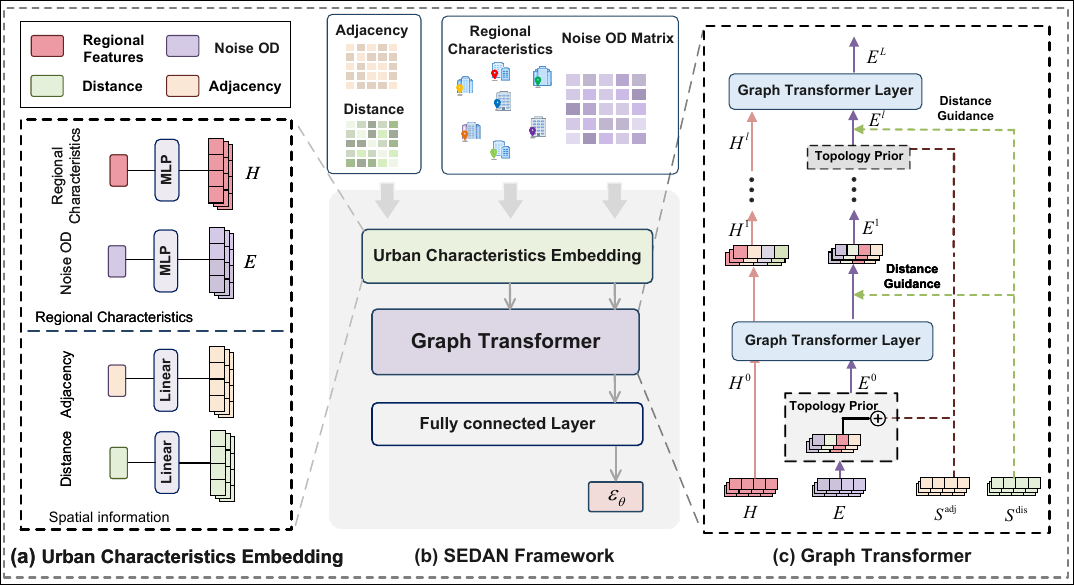}
\caption{Denoising network of SEDAN. 
(a) Urban characteristics embedding: regional attributes and the noisy OD matrix are mapped to initial node and edge features, and encoding spatial prior.
(b) Overall framework: regional attributes, a noisy OD matrix, and two spatial priors (adjacency and distance) are provided as inputs to the model. After the urban characteristics embedding, features pass through a stacked graph transformer and a fully connected layer to predict edge-level noise. 
(c) Graph Transformer: regional attributes serve as node inputs, and the noisy OD matrix provides edge inputs. Each layer jointly updates node and edge representations, where the topology prior preserves local connectivity and the distance guidance maintains global proximity. 
Color legend: regional features (red), noisy OD (purple), distance (green), adjacency (orange).}
\label{fig:denoising_network}
\end{figure}

During the reverse diffusion stage, the model progressively reconstructs the ground-truth OD matrix via the denoising network. To capture the joint effects of urban functional semantics and spatial structure on commuting flows, SEDAN leverages regional attributes to model interactions among urban regions, thereby characterizing latent commuting demand. Meanwhile, adjacency and distance information are injected as explicit spatial priors to guide the denoising process. The overall architecture is shown in \cref{fig:denoising_network}. 

The network takes the urban region feature matrix $\mathbf{X}_\mathcal{R}$ as node input and the noisy OD matrix $\mathbf{F}^{t}$ as edge input. We first embed regional features and the noisy OD matrix, and linearly project two spatial priors into learnable feature spaces:
\begin{equation}\label{eq:embed}
\begin{aligned}
\mathbf{H}^{\mathrm{0}} &= \mathrm{MLP}(\mathbf{X}_\mathcal{R}), &\qquad
\mathbf{E}^{\mathrm{0}} &= \mathrm{MLP}(\mathbf{F}^{t}),\\
\mathbf{S}^{\mathrm{adj}} &= \mathrm{Linear}(\mathbf{A}), &\qquad
\mathbf{S}^{\mathrm{dis}}  &= \mathrm{Linear}(\mathbf{D}),
\end{aligned}
\end{equation}
where $\mathbf{X}_\mathcal{R}\!\in\!\mathbb{R}^{N\times d_x}$ denotes attributes of $N$ regions (demographics, POIs), $\mathbf{F}^{t}\!\in\!\mathbb{R}^{N\times N}$ is the noisy OD matrix at timestep $t$. $\mathbf{A}\!\in\!\mathbb{R}^{N\times N}$ is the adjacency matrix and $\mathbf{D}\!\in\!\mathbb{R}^{N\times N}$ is the distance matrix.

To model how regions interact under spatial constraints, we stack graph transformer layers that iteratively update node and edge representations and output the edge-level noise $\epsilon_\theta$.  For the $k$-th head in layer $l$, the attention coefficient between node $i$ and its neighbor $j$ is computed as:

\begin{equation}\label{eq:attn}
\begin{gathered}
a_{ij}^{k,l} = \frac{Q^{k,l}h_i^l \cdot K^{k,l}h_j^l}{\sqrt{d_k}} + W^{k,l} \mathrm{ReLU}(\mathrm{Linear}([\,e_{ij}^{l};\,S_{ij}^{\mathrm{adj}}\,])),\\[2pt]
\alpha_{ij}^{k,l} = \mathrm{softmax}_{j}(a_{ij}^{k,l}),
\end{gathered}
\end{equation}
where $h_i^{l}$ and $e_{ij}^{l}$ are the node and edge features at the $l$-th layer, respectively. $Q^{k,l}$ and $K^{k,l}$ are the query and key matrices of the $k$-th attention head at the $l$-th layer. $W^{k,l}$ is the weight matrix of the $k$-th attention head at the $l$-th layer. $d_k$ is the dimension of the query and key vectors. The first term is the standard attention score, which captures semantic interactions between node features. The second term introduces a structure-aware bias by incorporating the adjacency prior into the edge representation, thereby encouraging stronger attention between geographically adjacent node pairs. As a result, the model preserves local connectivity while still capturing semantic dependencies from node features. Node representations are updated through the following formula:
\begin{equation}
h_i^{l+1} = O_h^l \parallel_{k=1}^{K} (\sum_{r_j \in \mathcal{N}_{r_i}}  \alpha_{ij}^{k,l} V^{k,l} h_j^l),
\end{equation}
where $V^{k,l}$ is the value matrix of the $k$-th attention head at the $l$-th layer. $O_h^{l}$ is the output MLP for the node features at the $l$-th layer. $K$ is the number of attention heads. $\mathcal{N}_{r_i}$ is the set of neighbor nodes that are connected to node $v_i$. To incorporate the distance prior, the attention logits from all heads are concatenated and projected into the edge feature space. The resulting edge representation is then combined with $S_{ij}^{\mathrm{dis}}$:

\begin{equation}
e_{ij}^{l+1} = O_e^l \parallel_{k=1}^{K} (a_{ij}^{k,l})+S_{ij}^{\mathrm{dis}},
\end{equation}
where $O_e^{l}$ is the output MLP for the edge features at the $l$-th layer.
After stacking $L$ layers, the edge-level noise is predicted by the fully connected layer:
\begin{equation}\label{eq:out}
\hat{\epsilon}_\theta=\mathrm{MLP}(\mathbf{E}^{L}\big).\qquad
\end{equation}

While spatial structure constraints enable the model to capture topological and geographical relationships, commuting OD flows still exhibit substantial numerical heterogeneity. This heterogeneity arises from variations in population density, infrastructure, and land use, frequently resulting in a long-tail distribution. Such skewness can impede model convergence and lead to unstable training. To alleviate this, we apply a logarithmic transformation to stabilize the value range:

\begin{equation}
\dot{F}_{ij} = \log(F_{ij}+1), \quad F_{ij} = \exp(\dot{F}_{ij})-1.
\end{equation}

The log-transformed flows $\dot{F}_{ij}$ are used during training, and recovered via exponential transformation during inference, mitigating the impact of extreme values.

\subsection{Model Training}
SEDAN adopts the standard loss formulation of diffusion probabilistic models to train the denoising network. Specifically, it minimizes the mean squared error between the predicted noise and the ground-truth noise added during the forward diffusion process. The loss function is defined as follows:
\begin{equation}
   \mathcal{L}={{\mathbb{E}}_{t,\epsilon \sim \mathcal{N}(0,I)}}\left[ \parallel \epsilon -{{\epsilon }_{\theta }}({\mathbf{F}^{t}},t,{\mathcal{C}_{\mathcal{R}}})\parallel _{2}^{2} \right].
\end{equation}
    
Although standard diffusion models generate high-quality samples, their iterative denoising steps lead to high computational overhead during inference. To improve efficiency, we adopt Denoising Diffusion Implicit Models (DDIM)~\cite{song2020denoising}, which accelerates sampling while preserving output fidelity.A detailed analysis of the inference time across different city scales is provided in ~\ref{sec:Inference_Time}.

The training procedure of the directed graph diffusion model is summarized in~\Cref{alg:training}. In each iteration, the model samples a directed graph from the training set and determines the diffusion step $t$. Gaussian noise is added to the clean OD matrix, and the denoising network predicts the injected noise given the noisy matrix, timestep, and urban features.   

\begin{algorithm}[ht]
\caption{Training Procedure of SEDAN}
\label{alg:training}
\begin{algorithmic}[1]
\Require A set of directed graphs $\mathcal{G}_{\text{train}}$,
\Ensure Trained network parameters $\theta$.

\State Sample a graph $G$ from $\mathcal{G}_{\text{train}}$ with OD matrix $\mathbf{F}$ and urban characteristics $\mathcal{C}_{\mathcal{R}}$
\State Sample timestep $t \sim \mathcal{U}\{1, 2, \ldots, T\}$
\State Sample noise $\epsilon \sim \mathcal{N}(0, \mathbf{I})$
\State Generate noisy input: $\tilde{\mathbf{F}}_t \leftarrow \sqrt{\bar{\alpha}_t}\,\mathbf{F} + \sqrt{1 - \bar{\alpha}_t}\cdot \epsilon$
\State Compute loss: $L \leftarrow \left\|\epsilon - \epsilon_{\theta}(\tilde{\mathbf{F}}_t, t, \mathcal{C}_{\mathcal{R}})\right\|_2^2$
\State Update parameters: $\theta \leftarrow \text{optimizer.step}(L)$
\end{algorithmic}
\end{algorithm}

After training, the model proceeds to the generation phase. As detailed in~\Cref{alg:generation}, we adopt the DDIM sampling strategy to efficiently generate OD matrices by reversing the diffusion process. Beginning with a pure Gaussian noise OD matrix, the model iteratively denoises it using the trained network, ultimately reconstructing an OD matrix that conforms to the spatial  characteristics  of the target city.  

\begin{algorithm}[ht]
\caption{Generation Procedure of SEDAN}
\label{alg:generation}
\begin{algorithmic}[1]
\Require urban characteristics $\mathcal{C}_{\mathcal{R}}$, 
         
         trained denoising network ${\theta}$, 
         
         DDIM sampling step size $\tau$,
\Ensure Generated OD matrix $\mathbf{F}$.

\State Initialize $\mathbf{F}^{T} \sim \mathcal{N}(0, \mathbf{I})$
\State Set step interval $\Delta t \leftarrow \frac{T}{\tau}$
\State \textbf{for} $t = T, T - \Delta t, \ldots, 1$ \textbf{do}
\State \quad $\mathbf{F}^{\,t-\Delta t} \leftarrow 
\frac{1}{\sqrt{\alpha_t}}
\left(
\mathbf{F}^{\,t}
- \frac{1-\alpha_t}{\sqrt{1-\bar{\alpha}_t}}\,
\epsilon_{\theta}(\mathbf{F}^{\,t},\, t,\, \mathcal{C}_{\mathcal{R}})
\right)$

\State \textbf{end for}
\State \textbf{return} $\mathbf{F}^{0}$
\end{algorithmic}
\end{algorithm}

\section{Experiment}

In this section, we first present the experimental settings, including dataset, model parameter configurations, evaluation metrics, and baseline methods. We then provide a detailed comparison of our approach with ten representative baselines. To evaluate the cross-city generalization capability of the proposed model, we perform experiments on cities of diverse scales and structural patterns. An ablation study further investigates the contribution of spatial constraints to OD flow generation. We further employ SHAP analysis to reveal key urban attributes that influence the generation process. Finally, a case study visualizes how the model leverages spatial priors, from local structural to global dependencies.

\subsection{Experimental Setup}

\subsubsection{Datasets}
This study utilizes the LargeCommuingOD dataset~\cite{ronglarge}, which covers 3,233 spatial units across the United States, including counties and census tracts. The dataset integrates multi-source information to construct regional features. Key data sources include the American Community Survey (ACS), the Longitudinal Employer-Household Dynamics (LODES) dataset, and OpenStreetMap (OSM). These sources collectively provide detailed descriptors of demographic composition and urban layout.

The LODES dataset constructs a national commuting OD matrix based on employer-employee relationships, reflecting commuting flows between home and workplace in 2018. The ACS provides 97 indicators spanning demographic and socioeconomic domains to characterize each region’s population composition and economic structure. Each entry in the OD matrix quantifies commuting flows from residential to workplace regions, characterizing the spatial structure of intercity mobility. To quantify spatial proximity and contiguity, the dataset includes a Euclidean distance matrix (computed from regional centroids) and an adjacency matrix (based on geographic boundaries).

\subsubsection{Evaluation Metrics}
The performance of the model is evaluated from two perspectives: numerical accuracy and spatial structural fidelity. 
For numerical accuracy, we employ Root Mean Square Error (RMSE)~\cite{liu2024spatial,liu2025st} and Normalized Root Mean Square Error (NRMSE)~\cite{rong2021inferring} to quantify the deviation between generated and ground-truth OD matrices. 
To assess spatial structural fidelity, we adopt two metrics: Common Part of Commuters (CPC)~\cite{liu2020learning} and Jensen–Shannon Divergence (JSD)~\cite{rong2023complexity,ronglarge}.
CPC measures the overlap between generated and real OD matrices in terms of OD pairs, with higher values indicating greater similarity in commuting patterns.
JSD evaluates the symmetric divergence between the probability distributions of generated and real data across three dimensions: inflow, outflow, and total OD flow. Lower JSD values suggest better alignment with actual spatial mobility structures, providing a measure of the model’s ability to reproduce realistic flow patterns.

All evaluation metrics are computed at the urban region level, and the final results are reported as the average across all cities. The specific formulas are defined as follows:

\begin{equation}
    \text{RMSE}=\sqrt{\frac{1}{|\mathbf{F}|}\underset{{{r}_{i}},{{r}_{j}}\in \mathcal{R}}{\mathop \sum }\,||{{F}_{ij}}-{{{\hat{F}}}_{ij}}||_{2}^{2}},
    \label{eq:RMSE}
\end{equation}
\begin{equation}
    \text{NRMSE}=\frac{\text{RMSE}}{\sqrt{\frac{1}{{{N}^{2}}}\mathop{\sum }_{{{r}_{i}},{{r}_{j}}\in \mathcal{R}}||{{F}_{ij}}-{{{\bar{F}}}_{ij}}||_{2}^{2}}} ,
    \label{eq:NRMSE}
\end{equation}
\begin{equation}
    \text{CPC}=2\cdot \frac{\mathop{\sum }_{{{r}_{i}},{{r}_{j}}\in \mathcal{R}}\min ({{F}_{i,j}},{{{\hat{F}}}_{i,j}})}{\mathop{\sum }_{{{r}_{i}},{{r}_{j}}\in \mathcal{R}}({{F}_{ij}}+{{{\hat{F}}}_{ij}})},
    \label{CPC}
\end{equation}
\begin{equation}
    \text{JSD}=\frac{\mathbf{KL}({{\mathbf{P}}_{\mathbf{F}}}||{{\mathbf{P}}_{{\hat{\mathbf{F}}}}})+\mathbf{KL}({{\mathbf{P}}_{{\hat{\mathbf{F}}}}}||{{\mathbf{P}}_{\mathbf{F}}})}{2},
    \label{eq:JSD}
\end{equation}
where $\bar{\mathbf{F}}$ denotes the mean of all elements in the ground-truth OD matrix $\mathbf{F}$, and $\mathbf{KL}$ refers to the Kullback–Leibler divergence. $\mathbf{P}$ represents the empirical probability distribution. The inflow of region $i$ is computed as the total volume received from all other regions $\sum_{j=1}^{N} F_{ji}$, while the outflow is defined as the total volume departing from region $\sum_{j=1}^{N} F_{ij}$.

\subsubsection{Baselines}
To systematically evaluate the performance of SEDAN, we conduct benchmark experiments across ten representative models, which fall into four major categories: physics-based models, data-driven models, element-wise prediction models, and matrix-level generative models. 

\textbf{(1) Physics-based models}

\begin{itemize}
  \item \textbf{GM-P}~\cite{zipf1946p}: Gravity Model with Power-law Decay. Assumes OD flow is proportional to the populations of the origin and destination, and inversely proportional to a power of the distance.
  \item \textbf{GM-E}~\cite{wilson1967statistical}: A variant of GM-P with exponential distance decay.
\end{itemize}

\textbf{(2) Statistical learning approaches}

\begin{itemize}
  \item \textbf{SVR}~\cite{rodriguez2021origin} : Support Vector Regression. An SVM-based regression that fits a maximum-margin function and ignores small errors to improve generalization.
  \item \textbf{RF}~\cite{pourebrahim2019trip,pourebrahim2018enhancing}: Random Forest. An ensemble of decision trees trained on bootstrap samples, with predictions aggregated by averaging. Captures complex non-linear dependencies in high-dimensional data.
  \item \textbf{GBRT}~\cite{robinson2018machine}: Gradient Boosting Regression Trees. Incrementally optimizes tree ensembles via gradient boosting, focusing on error minimization while adaptively controlling model complexity.
    \item \textbf{DGM}~\cite{simini2021deep}: Deep Gravity Model. Combines deep learning with traditional gravity formulations, learning flow distributions via neural networks to better capture high-dimensional features.
\end{itemize}

\textbf{(3) Element-wise prediction models}

\begin{itemize}
  \item \textbf{GMEL}~\cite{liu2020learning}: Leverages GNNs to integrate spatial and neighborhood information, enhancing regional embeddings via a multi-task framework and capturing interregional geographic dependencies for OD prediction.
\end{itemize}

\textbf{(4) Matrix-level generative models}

\begin{itemize}
  \item \textbf{NetGAN}~\cite{bojchevski2018netgan}: A GAN-based method that learns the distribution of random walks from real networks and generates new walks to reconstruct realistic graph topologies.
  \item \textbf{DiffODGen}~\cite{rong2023complexity}: A cascade diffusion model generating urban OD matrices, modeling graph topology and edge weights separately to improve large-scale prediction accuracy.
  \item \textbf{WeDAN}~\cite{ronglarge}: A denoising diffusion model generating OD matrices by learning regional attributes, effectively capturing spatial dependencies and outperforming traditional and deep generative baselines.
\end{itemize}

\subsubsection{Experiment Settings}
Experiments were conducted on a platform with an Intel Xeon Gold 6348 processor (2.60 GHz, 14 cores), 100 GB RAM, and a single NVIDIA A800 GPU (80 GB memory). We split the LargeCommuingOD dataset~\cite{ronglarge} into training, validation, and test sets in an 8:1:1 ratio. Each experiment was repeated five times independently, and the mean results were reported.

The diffusion model was trained with a 1000-step denoising process, using the cosine noise scheduling strategy proposed by Nichol et al.~\cite{nichol2021improved}. The denoising network was optimized using the AdamW optimizer~\cite{loshchilov2017decoupled} with a learning rate of 1e-4. SEDAN adopts a four-layer graph transformer architecture, with each layer consisting of 32 hidden units. During the generation phase, SEDAN, DiffODGen, and WEDAN each perform 10 stochastic sampling runs, and the mean of these samples is used as the final output. The gravity model follows the parameterization proposed by Barbosa et al.~\cite{barbosa2018human}, incorporating four tunable parameters. The random forest model is configured with 100 estimators. The Deep Generative Model (DGM)~\cite{simini2021deep} consists of 10 stacked layers, each with 64 hidden units. The GNN baseline is implemented with three layers, each containing 64 channels. The dual-cascade diffusion architecture and the denoising network configuration in DiffODGen~\cite{rong2023complexity} are kept consistent with those used in SEDAN. The hyperparameters of the model were selected based on validation set to balance predictive accuracy and computational efficiency. A sensitivity analysis is provided in Appendix~\ref{Hyperparameter_Analysis}.

\subsection{Comparisons with the SOTA Baselines}

To evaluate the modeling capacity and generalization performance of SEDAN, we perform a systematic comparison across ten representative models, with the results summarized in ~\Cref{tab:od_comparison}.

\begin{table}[htbp]
\centering
\caption{Performance Comparison on OD Matrix Generation. \textbf{Bold}: Best; \underline{Underline}: Second best.}
\label{tab:od_comparison}
\setlength{\tabcolsep}{3.8pt}
\begin{tabular}{@{}l p{2.7cm} ccc|cccc@{}}
\toprule
\multicolumn{2}{l}{\multirow{2}{*}{\textbf{Model}}} & \multicolumn{3}{c|}{\textbf{Flow Accuracy}} & \multicolumn{3}{c}{\textbf{Flow Distribution (JSD)}} \\
\cmidrule(lr){3-8}
\multicolumn{2}{l}{} & CPC↑ & RMSE↓ & NRMSE↓ & Inflow↓ & Outflow↓ & ODflow↓ \\
\midrule
& GM-P~\cite{zipf1946p}               & 0.32 & 174.00 & 2.22 & 0.67 & 0.66 & 0.41 \\
& GM-E~\cite{wilson1967statistical}   & 0.33 & 162.90 & 2.08 & 0.65 & 0.64 & 0.42 \\
\midrule
& SVR~\cite{rodriguez2021origin}       & 0.42 & 95.40  & 1.22 & 0.42 & 0.56 & 0.41 \\
& RF~\cite{pourebrahim2019trip,pourebrahim2018enhancing} & 0.46 & 100.40 & 1.28 & 0.42 & 0.50 & 0.22 \\
& GBRT~\cite{robinson2018machine}      & 0.46 & 91.00  & 1.62 & 0.42 & 0.49 & 0.23 \\
& DGM~\cite{simini2021deep}            & 0.43 & 92.90  & 1.19 & 0.47 & 0.56 & 0.23 \\
\midrule
& GMEL~\cite{liu2020learning}          & 0.44 & 94.30  & 1.20 & 0.45 & 0.36 & 0.21 \\
\midrule
& NetGAN~\cite{bojchevski2018netgan}   & 0.49 & 89.10  & 1.14 & 0.43 & 0.35 & 0.19 \\
& DiffODGen~\cite{rong2023complexity}  & 0.53 & 74.60  & 0.95 & 0.32 & 0.27 & 0.15 \\
& WEDAN~\cite{ronglarge}               & \underline{0.59} & \underline{68.60}  & \textbf{0.88} & \underline{0.29} & \underline{0.27} & \underline{0.15} \\
& \textbf{SEDAN} &
  \begin{tabular}[c]{@{}c@{}}\textbf{0.61}\end{tabular} &
  \begin{tabular}[c]{@{}c@{}}\textbf{63.54}\end{tabular} &
  \begin{tabular}[c]{@{}c@{}}\underline{0.93}\end{tabular} &
  \begin{tabular}[c]{@{}c@{}}\textbf{0.26}\end{tabular} &
  \begin{tabular}[c]{@{}c@{}}\textbf{0.21}\end{tabular} &
  \begin{tabular}[c]{@{}c@{}}\textbf{0.11}\end{tabular} \\
\bottomrule

\end{tabular}
\end{table}

The experimental results demonstrate that SEDAN consistently outperforms existing baseline methods across multiple key metrics. In terms of numerical accuracy, it improves the CPC score by 3.39\% compared to the strongest baseline, WEDAN. It also reduces RMSE by 7.38\%, indicating higher accuracy in generating commuting flows. The slight increase in NRMSE may result from SEDAN’s stronger emphasis on high-volume OD pairs. Although this trade-off leads to a minor rise in normalized errors, these OD pairs dominate the overall commuting network and offer greater practical value for urban traffic planning.
Regarding spatial distribution modeling, SEDAN significantly reduces the JSD for inflow, outflow, and OD flow by 10.34\%, 22.22\%, and 26.67\%, respectively. These gains reflect improved structural consistency between the generated and real-world OD matrices. Additionally, we draw the following findings:

\begin{itemize}
    \item \textbf{Data-driven models substantially outperform traditional physics-based models}. They capture complex nonlinear interregional interactions, enabling fine-grained human mobility modeling and preserving global spatial dependencies.
    \item \textbf{The joint OD matrix distribution learning strategy preserves spatial dependencies and global flow patterns} more effectively than pairwise modeling approaches (e.g., SVR, RF) that treat each OD pair independently.
    \item \textbf{Urban topological information enables strong cross-city generalization}, helping SEDAN maintain robust performance across diverse urban environments and consistently surpass existing graph-based generative models (e.g., NetGAN, DiffODGen).
\end{itemize}

\subsection{Analysis of Model Generalization Across Heterogeneous Urban Scenarios}

In commuting OD flow generation tasks, the spatial complexity of cities significantly impacts model performance. To systematically evaluate SEDAN's generalization ability, we categorize cities into three types based on the number of regions: small cities (0-10 regions), medium cities (11-50 regions), and large cities (51-100 regions). Based on this, we construct three test scenarios to progressively increase dataset heterogeneity: the low-heterogeneity scenario includes only small cities, the medium-heterogeneity scenario encompasses small and medium cities, and the high-heterogeneity scenario covers all three city types. We then compare SEDAN with the baseline model WEDAN to examine whether integrating spatial priors during the denoising process can maintain robust performance as dataset heterogeneity increases. The results are summarized in Table~\ref{tab:commuting_od_scale}.

\begin{table}[ht]
\centering
\caption{Performance Results Across Heterogeneous Urban Scenarios. \textbf{Bold}: Best.}
\label{tab:commuting_od_scale}
\small
\begin{tabular}{cll|cc}
\toprule
\textbf{Test Scenario} & \textbf{Category} & \textbf{Metric} & \textbf{WEDAN} & \textbf{SEDAN} \\
\midrule
\multirow{6}{*}{Low-Heterogeneity} 
  & \multirow{3}{*}{Flow Accuracy}
    & CPC↑     & 0.62 & \textbf{0.65} \\
  & & RMSE↓    & 83.16 & \textbf{74.40} \\
  & & NRMSE↓   & 1.03 & \textbf{0.96} \\
  \cline{2-5}
  & \multirow{3}{*}{Distribution (JSD)}
    & Inflow↓  & \textbf{0.34} & 0.35 \\
  & & Outflow↓ & 0.29 & \textbf{0.24} \\
  & & ODFlow↓ & 0.15 & \textbf{0.13} \\
\midrule
\multirow{6}{*}{Medium-Heterogeneity} 
  & \multirow{3}{*}{Flow Accuracy}
    & CPC↑     & 0.57 & \textbf{0.62} \\
  & & RMSE↓    & 71.48 & \textbf{66.79} \\
  & & NRMSE↓   & 1.04 & \textbf{0.97} \\
  \cline{2-5}
  & \multirow{3}{*}{Distribution (JSD)}
    & Inflow↓  & 0.30 & \textbf{0.28} \\
  & & Outflow↓ & 0.26 & \textbf{0.23} \\
  & & ODFlow↓ & 0.13 & \textbf{0.11} \\
\midrule
\multirow{6}{*}{High-Heterogeneity} 
  & \multirow{3}{*}{Flow Accuracy}
    & CPC↑     & 0.59 & \textbf{0.62} \\
  & & RMSE↓    & 72.96 & \textbf{63.47} \\
  & & NRMSE↓   & 1.04 & \textbf{0.93} \\
  \cline{2-5}
  & \multirow{3}{*}{Distribution (JSD)}
    & Inflow↓  & 0.31 & \textbf{0.26} \\
  & & Outflow↓ & 0.26 & \textbf{0.22} \\
  & & ODFlow↓ & 0.14 & \textbf{0.11} \\
\bottomrule
\end{tabular}
\end{table}

Although CPC slightly decreases for both models as urban heterogeneity
increases, SEDAN consistently outperforms WEDAN across all scenarios. Regarding error metrics, SEDAN achieves lower RMSE and NRMSE across all settings. From the low-heterogeneity to medium-heterogeneity scenarios, both models exhibit reduced RMSE values, indicating improved learning under moderate diversity. However, under high heterogeneity, WEDAN’s RMSE increases from 71.48 to 72.96, while SEDAN’s continues to decline, implying that SEDAN adapts better to complex and diverse urban environments.

In terms of distributional consistency, SEDAN also outperforms WEDAN across most JSD metrics. The only exception arises in the low-heterogeneity scenario, likely due to the limited spatial structure that constrains the effectiveness of its structure-guided design. In contrast, WEDAN, which relies less on spatial priors, generalizes slightly better under such low-structure conditions. As heterogeneity increases, SEDAN’s advantage in maintaining distributional alignment becomes increasingly evident. Based on these observations, the following conclusions can be drawn:
\begin{itemize}
\item \textbf{Model performance is affected by data heterogeneity.} As heterogeneity increases, WEDAN’s performance fluctuates, indicating that heterogeneous urban scenarios present challenges for models with limited spatial awareness.
\item \textbf{SEDAN remains robust with increasing urban heterogeneity and preserves spatial distributional consistency.} This suggests that leveraging regional semantics together with spatial priors helps the model adapt to city-specific mobility patterns while maintaining a stable spatial structure.
\end{itemize}

\subsection{Analysis of Model Robustness Across Cities with Different Structures}

To assess SEDAN’s adaptability to diverse urban forms, we develop a city structure classification method that considers both morphological and functional dimensions. The method identifies monocentric, uniform, and polycentric forms from four perspectives: population distribution, OD flows, commuting distance, and an integrated measure. The multi-perspective urban classification method is described in~\ref{sec:urban_structure_classification}.

\Cref{tab:classification} summarizes the proportion of cities in the LargeCommuingOD dataset~\cite{ronglarge} belonging to each structural type, based on our multi-perspective classification. Most cities are either polycentric or uniformly distributed in terms of population and OD flows, while commuting distances tend to follow a monocentric pattern. This heterogeneity may shape commuting behaviors and consequently influence the performance of OD flow generation models. We evaluate SEDAN in each structural category to assess its robustness and adaptability.

\begin{table}[htbp]
\centering
\caption{Urban Structural Typologies across Multiple Dimensions}
\label{tab:classification}
\begin{tabular}{lccc}
\toprule
\multirow{2}{*}{\textbf{Classification Dimension}} & \multicolumn{3}{c}{\textbf{Proportion of Cities}} \\
\cmidrule(lr){2-4}
 & \textbf{Monocentric} & \textbf{Uniform} & \textbf{Polycentric} \\
\midrule
Population Distribution & 23.80\% & 34.39\% & 41.81\% \\
OD Flow Distribution    & 29.00\% & 45.30\% & 25.70\% \\
Commuting Distance      & 45.12\% & 24.15\% & 30.73\% \\
Integrated Classification & 27.99\% & 30.73\% & 41.28\% \\
\bottomrule
\end{tabular}
\end{table}

\begin{figure}[ht]
  \centering
  \includegraphics[width=\textwidth]{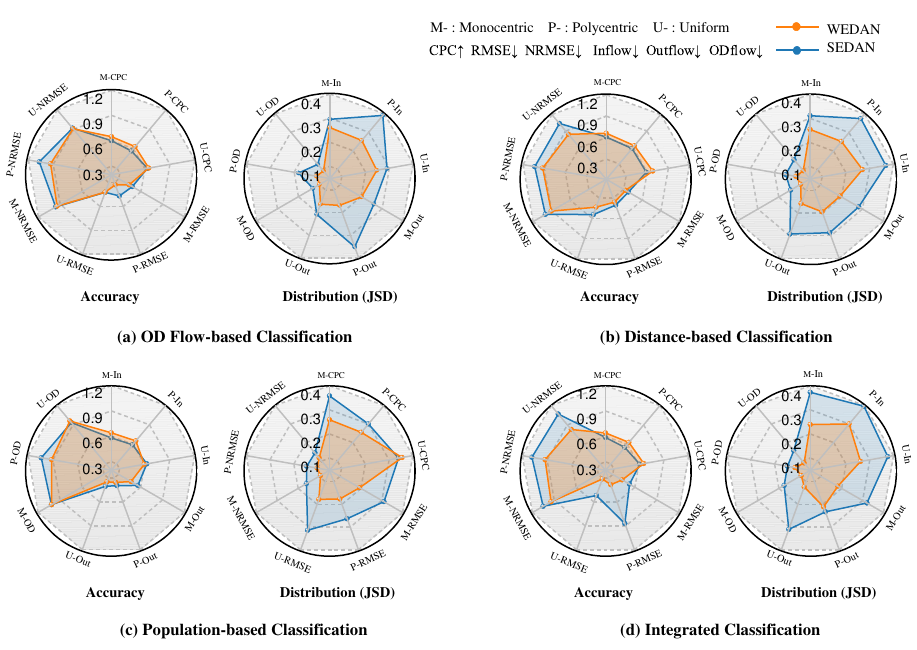} 
\caption{Performance across different urban structure. (a) OD flow–based classification, (b) distance–based classification, (c) population–based classification, and (d) integrated classification. 
Each subfigure contains two radar charts: the left summarizes 
\emph{accuracy} for cities grouped by structure—monocentric (M), polycentric (P), and uniform (U); the right reports \emph
{distributional consistency} measured by JSD for inflow, outflow, and OD flow. 
Orange lines denote WEDAN and blue lines denote SEDAN. 
For clarity, RMSE values are rescaled by a factor of $1/100$. This normalization does not affect the relative performance comparison.
}
  \label{fig:structure_radar}
\end{figure}

\cref{fig:structure_radar} presents the model performance across four classification perspectives---population distribution, OD flow, commuting distance, and the integrated perspective. The results of experiments indicate that SEDAN outperforms the baseline model WEDAN across most urban structure classifications, exhibiting notable improvements in both flow magnitude accuracy and spatial distribution consistency.

In the OD flow-based classification experiments, SEDAN consistently outperforms the baseline model WEDAN. Compared to WEDAN, SEDAN improves CPC by 8.0\% in monocentric cities and improves by 12.0\% in polycentric cities. It also significantly reduces both RMSE and NRMSE. The performance gains are especially notable in polycentric settings, where SEDAN models complex commuting patterns more effectively. In terms of distributional consistency, SEDAN achieves lower JSD values for inflow, outflow, and OD flow than WEDAN.

In the classification based on distance distribution, SEDAN also shows superior performance in terms of CPC and RMSE. Although the model shows a slight drop in CPC for uniformly structured cities under the population-based classification, it still performs well in monocentric and polycentric settings. This decline likely stems from the absence of distinct structural centers and dominant commuting flows in uniformly structured cities, where OD patterns are generally more diffuse and irregular. By contrast, in cities with pronounced spatial organization and complex inter-regional interactions, SEDAN is able to more effectively model mobility patterns by jointly leveraging spatial priors and regional semantics.

By jointly considering OD flows, spatial distance, and population distribution, SEDAN achieves improvements over WEDAN across all evaluation metrics. Particularly in polycentric urban structures, it achieves a 13.9\% higher CPC, reduces RMSE from 86.90 to 38.03, and lowers OD flow JSD from 0.146 to 0.114. These results highlight SEDAN’s robustness in modeling complex urban mobility patterns.

This robustness is rooted in SEDAN’s effective fusion of spatial structure and regional semantics. Spatial priors provide a structural foundation that delineates geographically plausible connections, while regional attributes supply functional context, enabling the model to identify flows that reflect actual mobility demand. Through this integrated modeling, SEDAN captures universal mobility patterns, leading to consistent performance across monocentric, uniform, and polycentric cities.

\subsection{Feature Missing Robustness Analysis}

In cross-city OD generation, the target city, especially an underdeveloped city, often lacks complete fine-grained POI or demographic data. To evaluate model robustness under such conditions, we conduct a feature-missing experiment. Specifically, 10\%, 30\%, and 50\% of node features are randomly masked, and missing values are imputed using the city-level mean of each feature. The results are reported in Table~\ref{tab:mask_ratio_results}.

As the masking ratio increases, the performance of both WEDAN and SEDAN declines, but SEDAN consistently outperforms WEDAN across all settings. The CPC of SEDAN remains above 0.56 under all masking ratios, whereas that of WEDAN drops from 0.56 to 0.53. Even under the severe 50\% masking condition, SEDAN still achieves a lower RMSE (69.45) than WEDAN (76.19), indicating stronger robustness when semantic features are heavily corrupted.

SEDAN also shows better distributional consistency. Under 50\% masking, its ODFlow JSD is 0.14, compared with 0.19 for WEDAN. These results suggest that adjacency and distance priors provide stable spatial structural constraints when semantic features are degraded, thereby partially compensating for the missing information.

\begin{table}[h]
\centering
\caption{Performance under Varying Feature Mask Ratios. \textbf{Bold}: Best.}
\label{tab:mask_ratio_results}
\small
\begin{tabular}{cll|cc}
\toprule
\textbf{Mask Ratio} & \textbf{Category} & \textbf{Metric} & \textbf{WEDAN} & \textbf{SEDAN} \\
\midrule
\multirow{6}{*}{10\%} 
  & \multirow{3}{*}{Flow Accuracy}
    & CPC$\uparrow$     & 0.56 & \textbf{0.61} \\
  & & RMSE$\downarrow$    & 72.37 & \textbf{64.69} \\
  & & NRMSE$\downarrow$   & 1.05 & \textbf{0.95} \\
  \cline{2-5}
  & \multirow{3}{*}{Distribution (JSD)}
    & Inflow$\downarrow$  & 0.33 & \textbf{0.31} \\
  & & Outflow$\downarrow$ & 0.28 & \textbf{0.22} \\
  & & ODFlow$\downarrow$  & 0.16 & \textbf{0.12} \\
\midrule
\multirow{6}{*}{30\%} 
  & \multirow{3}{*}{Flow Accuracy}
    & CPC$\uparrow$     & 0.55 & \textbf{0.59} \\
  & & RMSE$\downarrow$    & 74.14 & \textbf{67.24} \\
  & & NRMSE$\downarrow$   & 1.14 & \textbf{0.98} \\
  \cline{2-5}
  & \multirow{3}{*}{Distribution (JSD)}
    & Inflow$\downarrow$  & 0.34 & \textbf{0.32} \\
  & & Outflow$\downarrow$ & 0.31 & \textbf{0.24} \\
  & & ODFlow$\downarrow$  & 0.17 & \textbf{0.13} \\
\midrule
\multirow{6}{*}{50\%} 
  & \multirow{3}{*}{Flow Accuracy}
    & CPC$\uparrow$     & 0.53 & \textbf{0.56} \\
  & & RMSE$\downarrow$    & 76.19 & \textbf{69.45} \\
  & & NRMSE$\downarrow$   & 1.15 & \textbf{1.01} \\
  \cline{2-5}
  & \multirow{3}{*}{Distribution (JSD)}
    & Inflow$\downarrow$  & 0.37 & \textbf{0.33} \\
  & & Outflow$\downarrow$ & 0.32 & \textbf{0.27} \\
  & & ODFlow$\downarrow$  & 0.19 & \textbf{0.14} \\
\bottomrule
\end{tabular}
\end{table}

\subsection{Ablation Study}

To systematically evaluate the contribution of each component in SEDAN, we conduct an ablation study from four aspects: spatial priors, semantic information (demographics and POIs), fusion mechanism, and the diffusion generation mechanism. We consider six variant models:
\begin{itemize}
    \item \textbf{w/o A:} The adjacency matrix is removed, preventing the model from explicitly leveraging topological relationships among regions.
    \item \textbf{w/o D:} The distance matrix is removed, eliminating information on spatial proximity and travel resistance.
    \item \textbf{w/o AD:} Both adjacency and distance matrices are removed, leaving the model to rely solely on regional features.
    \item \textbf{w/o POI:} The POI features are removed.
    \item \textbf{w/o Diffusion:} The diffusion process is removed, and OD flows are directly predicted through the Graph Transformer.
    \item \textbf{w/o Constraint:} The proposed fusion mechanism is removed, and spatial priors are simply treated as edge features.
\end{itemize}

\begin{figure}[ht]
\centering
\includegraphics[width=\linewidth]{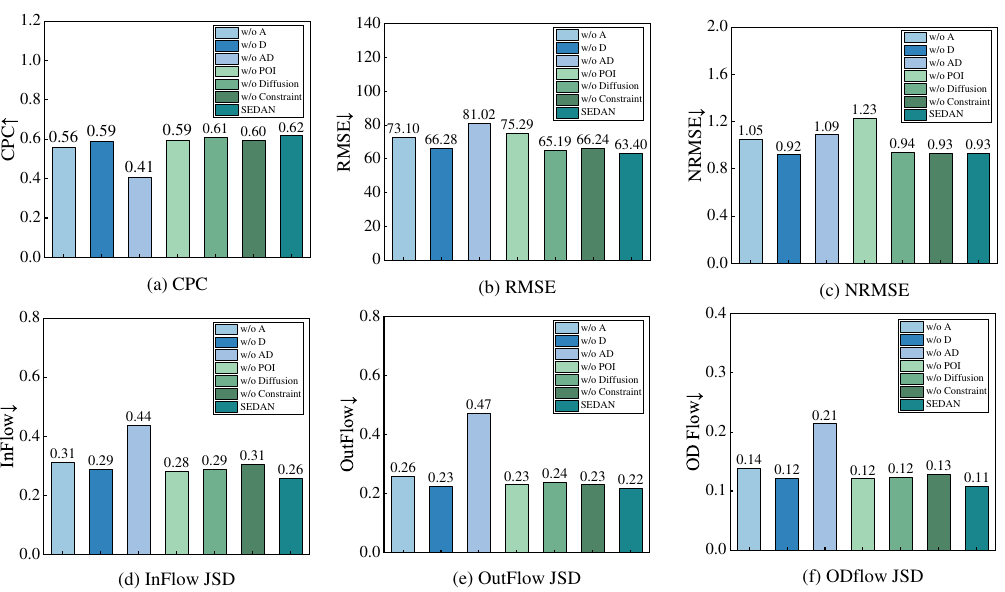}
\caption{Ablation study results. Comparison of the full SEDAN model against six variants: 
\textbf{w/o A} (without adjacency prior), \textbf{w/o D} (without distance prior), 
\textbf{w/o AD} (without either prior), \textbf{w/o POI} (without POI features), 
\textbf{w/o Diffusion} (without the diffusion process), and \textbf{w/o Constraint} 
(without spatial information fusion mechanism). 
Performance is reported on six metrics: (a) CPC$\uparrow$, (b) RMSE$\downarrow$, 
(c) NRMSE$\downarrow$, (d) Inflow JSD$\downarrow$, (e) Outflow JSD$\downarrow$, 
and (f) ODFlow JSD$\downarrow$.}
\label{fig:ablation}
\end{figure}

From the perspective of spatial priors, w/o AD performs the worst on almost all metrics and shows the largest increase in JSD, indicating a substantial discrepancy between the generated and real OD distributions. Removing either the adjacency matrix or the distance matrix also degrades performance, suggesting that the two types of spatial priors play complementary roles in the diffusion process. In addition, although w/o Constraint still uses complete spatial priors as inputs, it remains inferior to the full model. This suggests that performance depends not only on the availability of spatial priors, but also on their integration into the model.

From the semantic perspective, removing POI features leads to clear performance degradation, especially in numerical error metrics: the RMSE increases from 63.40 to 75.29, and the NRMSE rises from 0.93 to 1.23. This suggests that POI information, which reflects regional functional semantics, is crucial for capturing urban commuting demand.

Removing the diffusion process worsens distribution-related metrics: the Inflow JSD increases from 0.26 to 0.29, and the Outflow JSD rises from 0.22 to 0.24. This suggests that while direct regression fits flow intensity well, it is less effective at recovering the global flow distribution. In contrast, diffusion better captures the underlying data distribution. Although diffusion introduces additional computational cost, it is acceptable for offline applications such as transportation planning~\cite{zeng2022causal} and public health modeling~\cite{lu2026human}, where distribution fidelity matters more than inference efficiency. These results lead to three main findings:
\begin{itemize}
    \item Semantic and spatial information play distinct but complementary roles: semantic features capture regional functions and latent travel demand, whereas spatial priors provide structural constraints for OD generation.
    \item Transforming spatial priors into explicit structural constraints is more effective than simply treating them as input features.
    \item Adjacency and distance information are highly complementary, enabling the model to preserve both local topology and global spatial regularity.
\end{itemize}

\subsection{Feature Contribution Analysis}

The generation of OD matrices is a complex task, as commuting flows are influenced by a variety of factors in different ways. Identifying key features and understanding their contributions are crucial for improving model performance and interpretability. We employ SHAP (Shapley Additive Explanations)~\cite{lundberg2017unified}  to quantify the relative importance of different features. The explanation target is defined as the average outflow of a target region. We use KernelSHAP to estimate the marginal contribution of each input feature. The absolute SHAP value reflects the feature's importance, with larger values indicating a greater contribution to the model's output. Details are provided in~\ref{sec:shap}.

The OD matrix is derived from the LODES dataset, which records home-to-work commuting flows based on employer–employee relationships. Therefore, the features analyzed in this section describe regional job–housing structure at the macro level.

\begin{figure}[H]
\centering
\includegraphics[width=1\textwidth]
{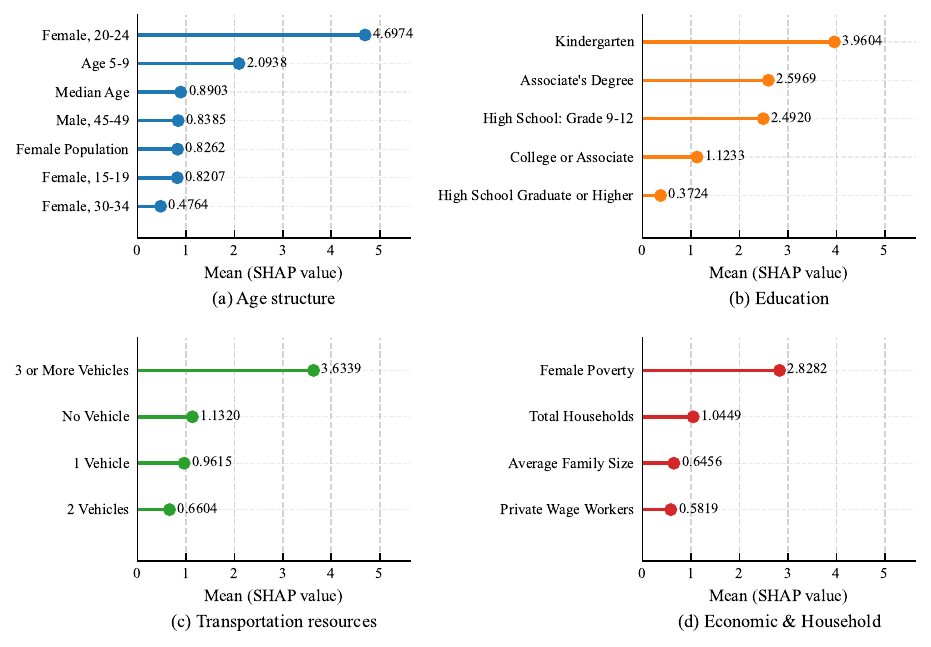}
\caption{SHAP values of the top 20 demographic features. The selected features are grouped into four dimensions: (a) age structure, (b) education, (c) transportation resources, and (d) economic and household characteristics.}
\label{fig:shap_demo}
\end{figure}

Figure~\ref{fig:shap_demo} presents the SHAP values of the top 20 demographic features, which can be grouped into four dimensions: age structure, education, transportation resources, and economic and household characteristics. Overall, age-related features show the strongest effects, while education and transportation resources also contribute substantially. Economic and household characteristics have relatively weaker effects.

Among the age-related features, Female, 20–24 has the highest importance. This may reflect the prominent role of younger populations in commuting, as well as the tendency for women to work in sectors with more regular commuting patterns, such as education and healthcare. "Age 5-9" and "Female, 15-19" are also important. These variables may indirectly reflect household structure. Areas with more school-age children often indicate a higher concentration of young families and thus stronger commuting demand. "Median Age", "Male, 45-49", and "Female, 30-34" reflect the stable commuting demand of the middle-aged workforce.

Education-related features cover several stages, from preschool to secondary and higher education. Population at basic education stages often reflects the concentration of young families and the associated commuting demand. Populations with higher education levels directly represent the potential commuting population.

Transportation resources also play an important role. "3 or More Vehicles" shows the strongest effect, as greater vehicle availability usually corresponds to higher travel flexibility. Meanwhile, "No Vehicle" also shows relatively high importance, suggesting that these populations rely more heavily on public transportation.  

In the economic and household dimension, "Female Poverty" shows the strongest effect, possibly reflecting livelihood-related travel under economic constraints. "Total Households" and "Average Family Size" capture the underlying scale of commuting demand.

\begin{figure}[H]
\centering
\includegraphics{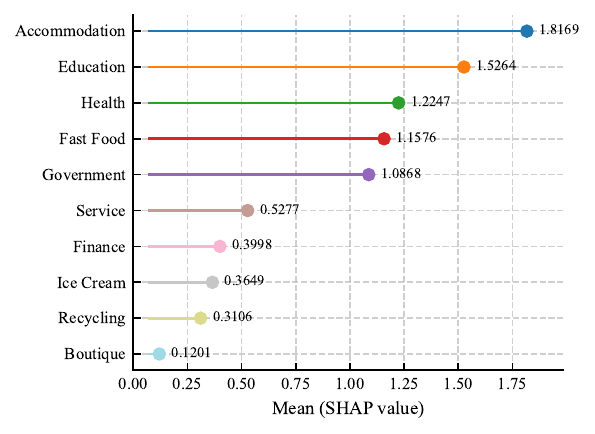}
\caption{SHAP values of the top 10 POI features}
\label{fig:shap_poi}
\end{figure}

Figure~\ref{fig:shap_poi} shows the SHAP values of the top 10 POI features. These features reflect urban functional characteristics. Among them, "Accommodation" has the highest importance, suggesting that residential areas are the main origins of commuting flows. "Education", "Health", and "Government" are also highly important. These POIs are often associated with urban centers and therefore help shape commuting patterns. "Fast Food" also shows a notable effect, as such facilities are often located in busy urban areas. By contrast, "Ice Cream", "Recycling", and "Boutique" are non-core amenities, and their effects on commuting flows are relatively limited.

Overall, SHAP analysis reveals the critical influence of demographic and POI features in the OD flow generation, indicating that model predictions are jointly driven by socioeconomic characteristics and spatial environmental factors. Population composition, transportation accessibility, and the spatial layout of functional facilities collectively play a central role in shaping OD distribution.

\subsection{Case Study}
\label{subsec:attn_case_study}

We conduct a case study to examine the attention patterns across layers and evaluate how the model leverages spatial priors. By comparing the attention matrices with the adjacency and distance-based similarity matrices, we analyze how SEDAN captures spatial dependencies at different stages.

\begin{figure}[t]
    \centering
    \includegraphics[width=\linewidth]{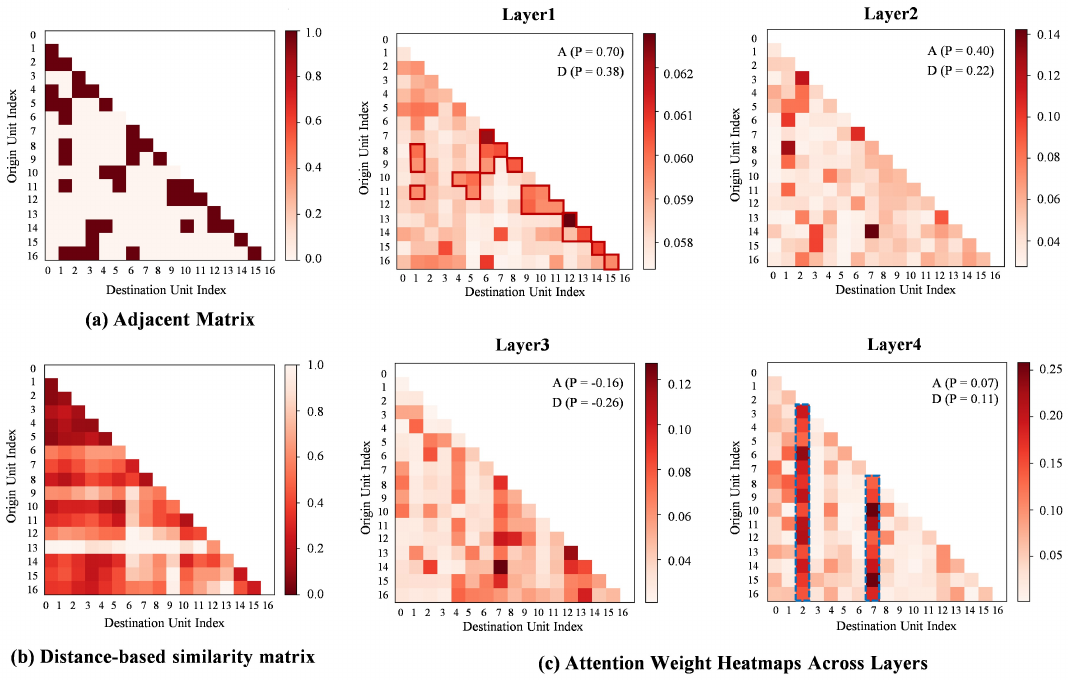}
    \caption{
    Visualization of attention weights across transformer layers.
    (a) Binary adjacency matrix representing the graph topology.
    (b) Distance-based similarity matrix derived from physical distance, where shorter distances correspond to higher similarity. 
    (c) Attention weight heatmaps across four transformer layers. Darker colors indicate larger attention weights between origin--destination region pairs.
    Each attention map reports its Pearson correlation (P) with the adjacency matrix (A) and the distance-based similarity matrix (D), computed over off-diagonal elements. A higher positive correlation indicates that the attention pattern is more consistent with the corresponding spatial prior, whereas a lower or negative correlation suggests weaker dependence on spatial structure.
    }

    \label{fig:attn_case_study}
\end{figure}

As shown in~\cref{fig:attn_case_study}, the shallow layers rely heavily on spatial information. In Layer~1, several high-attention regions closely match the adjacency matrix, as highlighted by the red solid boxes. This observation is consistent with the strong correlation with adjacency (P = 0.70), suggesting that the model initially aggregates information mainly from nearby regions.
This structural reliance begins to decline in Layer~2, where both correlations drop markedly, suggesting a shift toward cross-region interactions. By Layer~3, the correlations become negative, implying that the model is exploring non-local or long-range dependencies. In Layer~4, strong attention from multiple origins converges on two destination regions, as highlighted by the blue dashed boxes. This pattern suggests that the model has identified certain nodes as potential hubs or bridges that play a central role in information flow.

These results reveal a clear hierarchical learning strategy. The shallow layers are primarily guided by spatial priors, grounding the model in the underlying spatial structure. In the intermediate layers, these constraints are gradually relaxed, allowing the model to capture interregional mobility patterns shaped by functional and demographic characteristics. The deep layers reflect a deeper integration of spatial and regional attributes, enabling the model to identify key roles such as hubs and bridges. This progression illustrates how SEDAN effectively combines spatial structure with regional attributes to build a robust understanding of mobility patterns.

\section{Discussion}
By integrating regional features and spatial priors into a conditional diffusion framework, SEDAN effectively addresses the challenges of generating accurate OD flows and generalizing across heterogeneous urban settings. The results convey three main messages: (1) SEDAN explicitly fuses regional features and spatial structure priors during the diffusion process, enabling unified modeling of node semantics and spatial structure. This improves both the accuracy and structural fidelity of the generated OD matrices. (2) The performance gain is especially pronounced in polycentric cities, suggesting that structural priors help mitigate structural drift and enhance robustness under complex urban structure. (3) The adjacency matrix provides local connectivity, while the distance matrix captures global proximity relations. These priors impose complementary local-to-global constraints, improving the distributional consistency of the generated OD matrices. 

\noindent\textbf{\textit{Drawbacks.}} Despite the strong results, this work has several limitations. (1) \emph{Lack of temporal dynamics}: The model is trained on static OD data and does not capture temporal evolution patterns influenced by holidays, special events, or emergencies. (2) \emph{Simplified spatial priors}: The current priors are limited to adjacency and distance matrices. Richer sources such as road networks and public transit are not yet incorporated. (3)\emph{Assumption of feature-space consistency}: The model assumes consistent semantics for features such as population and facilities across cities. However, differences in statistical standards between cities may introduce mismatches, which can degrade model stability in highly heterogeneous settings. (4)\emph{Limited regional diversity}: This study is based on data from U.S. cities. While the results are encouraging, the applicability of the model to cities in other geographic and socioeconomic contexts remains worth exploring.

\noindent\textbf{\textit{Potential Avenues.}} To address these limitations, future research may explore the following directions. (1)\emph{Temporal dynamics}. Introduce time series modeling to capture the evolution of OD flows, enabling the model to capture temporal dependencies. (2)\emph{Multi-view spatial fusion}: Build a multi-view modeling framework that integrates road network, functional zoning, and transport infrastructure to better capture urban spatial relationships and mobility patterns. (3) \emph{City similarity constraints}: Incorporate inter-city similarity as a guidance within diffusion to promote pattern transfer from structurally similar cities and enhance cross-city generalization. (4)\emph{Learning shared mobility regularities}: Future work could investigate how to learn mobility regularities shared across cities while preserving city-specific characteristics, thereby improving adaptability to cities from different countries.

\noindent\textbf{\textit{Implications.}} Methodologically, SEDAN shows that injecting explicit structural priors into diffusion models improves both accuracy and structural realism. Practically, these improvements are also meaningful for real-world applications. High-fidelity OD matrix generation can support a variety of downstream tasks, including urban planning~\cite{imai2021origin,zeng2022causal,credit2023method}, transportation carbon emission assessment~\cite{zeng2024estimating}, and public health modeling~\cite{mistry2021inferring,lu2026human}. More broadly, SEDAN underscores the value of coupling deep representations with structural priors when handling large-scale, heterogeneous urban data. This structure-guided diffusion paradigm is not confined to urban systems and may also be applicable to other complex networks, such as logistics systems, epidemic propagation, and information diffusion.
\section{Conclusion}
This study proposes SEDAN, a structure-enhanced diffusion model for OD matrix generation that simultaneously achieves accurate flow generation and robust generalization across heterogeneous cities. The core design of SEDAN lies in formulating OD matrix generation as a task driven by the coupling of multi-source heterogeneous information, jointly modeling urban semantics and spatial structure. The model incorporates two complementary spatial priors into the graph transformer: the adjacency matrix preserves local topological connectivity, while the distance matrix captures global spatial proximity. Under these spatial constraints, SEDAN models how functional and demographic characteristics jointly determine latent commuting demand. Via this deep fusion of semantic and structural information, SEDAN captures complex nonlinear inter-regional interactions and generates OD matrices that are both geographically consistent and behaviorally meaningful. Experimental results demonstrate that SEDAN outperforms the SOTA baseline, WEDAN: the JSD of inflow, outflow, and overall OD flow decreases by 10.34\%, 22.22\%, and 26.67\%.  SEDAN provides a generalizable framework for modeling complex urban mobility patterns and can be extended to other spatial graph generation tasks.

\bibliographystyle{elsarticle-num}
\bibliography{references}

\newcounter{saveeqn}
\setcounter{saveeqn}{\value{equation}}

\appendix

\counterwithout{equation}{section}

\setcounter{equation}{\value{saveeqn}}
\renewcommand{\theequation}{\arabic{equation}}

\appendix

\section{Analysis of Spatial Regularities in Commuting Flows}
\label{sec:Spatial_Regularities}

To uncover the underlying patterns of commuting behaviors, we analyzed the OD data across all cities in the dataset. Table~\ref{tab:global_stats} summarizes these statistics. The results reveal a pronounced distance-decay effect, evidenced by a strong negative correlation ($-0.472$) between distance and log-transformed flows. Furthermore, geographical proximity plays a crucial role, with adjacent regions exhibiting notably higher flow intensity and a lower prevalence of zero-flow pairs.

\begin{table}[h]
\centering
\caption{spatial statistics of commuting flows.}
\label{tab:global_stats}
\begin{tabular}{lc}
\toprule
\textbf{Metric} & \textbf{Value} \\
\midrule
Distance-LogFlow Correlation & $-0.47$ \\
Mean Flow (Adjacent OD Pairs) & $80.00$ \\
Mean Flow (Non-adjacent OD Pairs) & $36.28$ \\
Non-zero Rate (Adjacent OD Pairs) & $96.79\%$ \\
Non-zero Rate (Non-Adjacent OD Pairs) & $88.31\%$ \\
\bottomrule
\end{tabular}
\end{table}

\begin{figure}[h]
    \centering
    \includegraphics[width=0.60\linewidth]{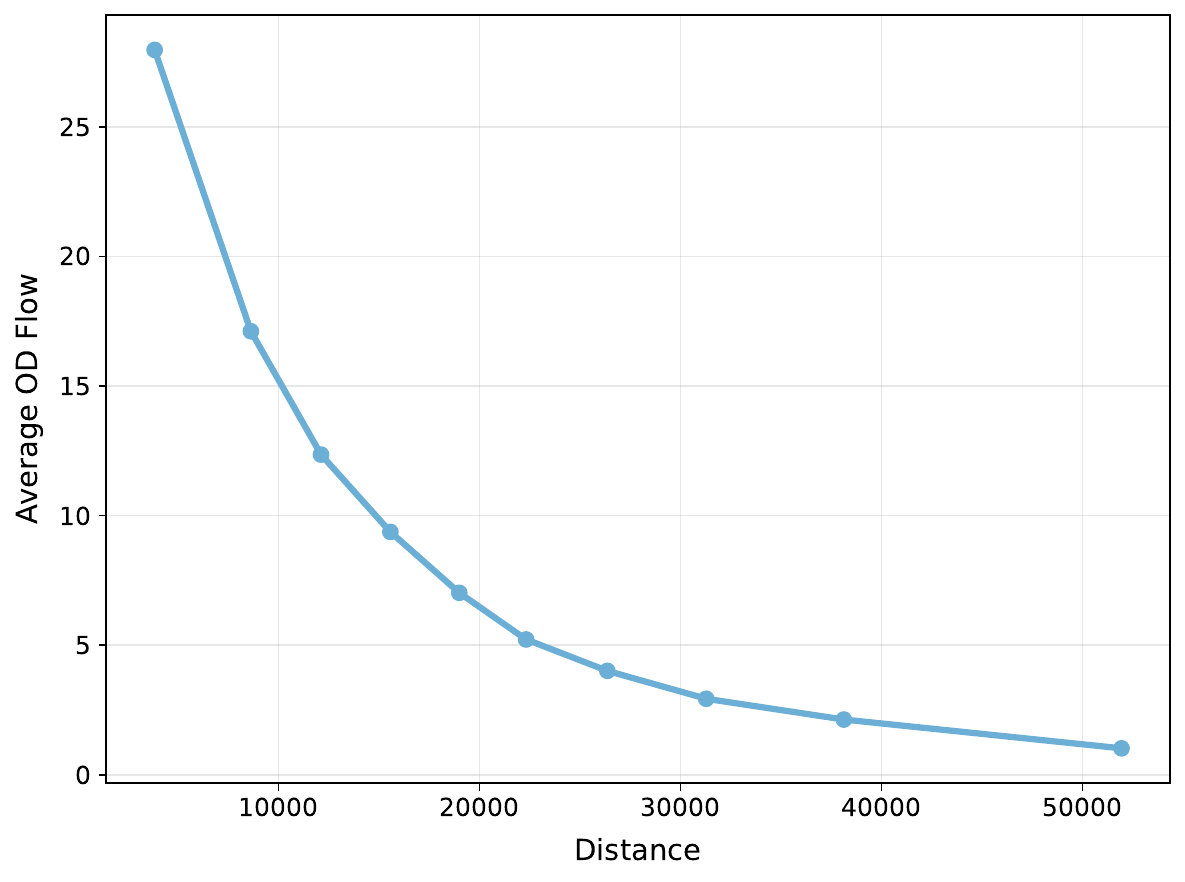}
    \caption{Distance-decay pattern of OD flows.}
    \label{fig:distance_decay}
\end{figure}

\begin{figure}[htpb]
    \centering
    \includegraphics[width=0.8\linewidth]{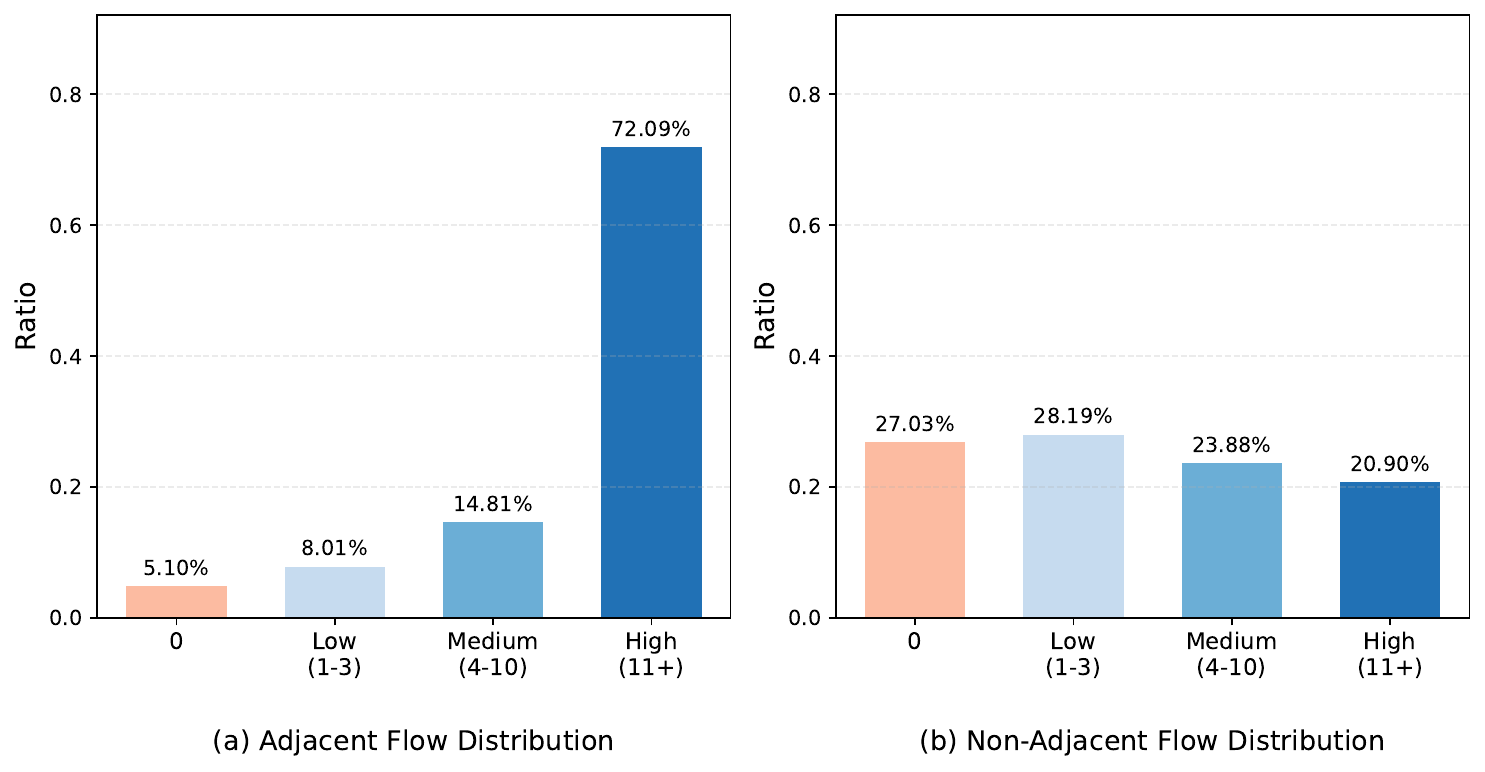}
    \caption{The distribution of OD flows. (a) Distribution for adjacent region pairs; (b) Distribution for non-adjacent region pairs.}
    \label{fig:adj_dist}
\end{figure}

These findings reveal spatial regularities across cities. At the city level, such patterns also evident. As illustrated in Figure~\ref{fig:distance_decay}, the average OD flow decreases with increasing spatial distance, forming a clear distance-decay trend. In addition, Figure~\ref{fig:adj_dist} shows that high-flow interactions are predominantly concentrated among adjacent regions, whereas non-adjacent pairs are largely associated with zero or low-volume flows. In summary, although urban characteristics vary widely, spatial regularities rooted in distance and topological adjacency demonstrate a high degree of consistency across different cities.

\section{Urban Center Structure Classification Method}
\label{sec:urban_structure_classification}
The structure of urban centers provides a fundamental basis for understanding the spatial organization and the distribution of job–housing relationships within cities. Prior studies suggest that urban structure can be characterized from both morphological and functional perspectives~\cite{zhu2024polycentric,tang2022exploring}. To systematically identify center structure types across different cities, we propose a comprehensive evaluation method that integrates both dimensions. This method provides a solid foundation for evaluating model generalization across structurally diverse cities.

This method constructs a multi-dimensional indicator system from three perspectives: population distribution, OD flow, and commuting distance. It is designed to characterize the spatial organization of urban areas and the patterns of job–housing interaction, particularly in terms of their concentration and spatial dispersion.
\begin{itemize}
    \item Population dimension: Includes the Pareto index, primacy index, Gini coefficient, and Herfindahl index, which respectively measure the power-law property of population distribution, dominance of core areas, overall balance of population distribution, and concentration level of population.
    
    \item OD flow dimension: Employs the Gini coefficient of flows, Herfindahl index, maximum flow proportion, and maximum betweenness centrality to characterize the inequality of flows, concentration of dominant nodes, and the influence of key commuting hubs.

    \item Commuting distance dimension: Utilizes the same four indicators as the OD flow dimension to quantify the spatial dispersion and centrality of commuting behaviors.
\end{itemize}

These indicators collectively reflect both static morphological characteristics and dynamic functional interactions. To integrate the diverse measures, all indicators are first standardized, followed by Principal Component Analysis (PCA) to extract key components and compute scores for each dimension. These scores are then weighted and aggregated into a composite structural score for each city. Subsequently, K-means clustering is employed to classify cities into three structural types: monocentric, uniform, and polycentric. The definitions of the metrics are presented as follows:

\textbf{Gini Coefficient}:  
Measures the level of imbalance in data distribution. It evaluates how evenly population, commuting volume, and commuting distance are distributed across a city. A higher Gini value indicates greater imbalance, suggesting a tendency toward monocentric urban structure. It is computed as:
\begin{equation}
\text{G} = \frac{\sum_{i=1}^{N} \sum_{j=1}^{N} |X_i - X_j|}{2N^2 \bar{X}},
\label{eq:gini}
\end{equation}
where $X_i$ represents the value of region $i$ in a given dimension (e.g., population, commuting volume, or commuting distance), $\bar{X}$ is the average value across all regions, and $N$ is the total number of regions. A Gini coefficient approaching 0 indicates a relatively balanced distribution, implying a polycentric urban structure, while values approaching 1 indicate high concentration, reflecting a monocentric form.

\textbf{Herfindahl-Hirschman Index (HHI)}:  
A key measure of concentration used to assess spatial clustering of urban functions. Values closer to 1 indicate a more monocentric pattern. It can be expressed as:
\begin{equation}
\label{eq:hhi}
\text{HHI} = \sum_{i=1}^{N} \left( \frac{X_i}{\sum_{j=1}^{N} X_j} \right)^2,
\end{equation}
where $X_i$ represents the value of region $i$ in a given dimensio (e.g., population, commuting volume, or commuting distance) and $N$ is the total number of regions. HHI close to 1 indicates concentration in a few regions (monocentric), while values near 0 suggest polycentricity.

\textbf{Maximum Flow Share Ratio (MFS)}:  
Measures the dominance of the largest commuting volume or distance in the city. It is defined as:
\begin{equation}
\label{eq:mfs}
\text{MFS} = \frac{\max (X_i)}{\sum_{j=1}^{N} X_j},
\end{equation}
where $\max(X_i)$ is the maximum commuting volume or distance, and $\sum_{j=1}^{N} X_j$ is the corresponding total of that variable. Higher values indicate monocentricity, whereas lower values indicate polycentricity.

\textbf{Maximum Betweenness Centrality (MBC)}:  
Assesses the importance of key nodes in the urban network. It is formulated as:
\begin{equation}
\label{eq:mbc}
\text{MBC} = \max \left( \sum_{s \ne i \ne t} \frac{\sigma_{st}(i)}{\sigma_{st}} \right),
\end{equation}
where $\sigma_{st}$ is the total number of shortest paths between $s$ and $t$, and $\sigma_{st}(i)$ is the count passing through node $i$. Higher values indicate stronger monocentricity.

\textbf{Pareto Exponent}:  
Describes the degree of imbalance under the assumption of a power-law distribution. It is calculated as:
\begin{equation}
\label{eq:pareto}
\alpha = \frac{1}{1 - \frac{\sum_{i=1}^N \log P_i}{N \log P_{\max}}},
\end{equation}
where $P_i$ is the population of region $i$, and $P_{\max}$ is the population of the largest region. Higher $\alpha$ indicates monocentricity, lower $\alpha$ suggests polycentricity.

\textbf{Primacy Index:}
The Primacy Ratio is an important indicator used to measure the relative dominance of the largest region within a city. It is calculated as the ratio of the population in the largest region to the combined population of the second, third, and fourth largest regions:

\begin{equation}
\label{eq:primacy}
\text{PI} = \frac{P_1}{P_2 + P_3 + P_4},
\end{equation}
where $P_1$ denotes the population of the largest region, and $P_2$, $P_3$, and $P_4$ represent the populations of the next three largest regions, respectively. Higher PI indicates a highly centralized urban structure dominated by a single core region. In contrast, lower PI suggests a more balanced population distribution and a tendency toward a polycentric urban form.

\section{Inference Time Analysis Under Different City Scales}\label{sec:Inference_Time}

Diffusion models rely on multi-step iterative sampling, which usually leads to higher inference overhead. To evaluate the practicality of the proposed method in large-scale urban scenarios, we conduct a quantitative analysis of inference time under different city scales, represented by the number of nodes $N$.

Specifically, we compare the average inference time of three models: SEDAN (the proposed method), WEDAN (a representative diffusion-based baseline), and GMEL (a representative regression-based model). For the diffusion models SEDAN and WEDAN, accelerated sampling is performed using DDIM, with the number of sampling steps set to $K=100$. In the experiments, we record the average inference time required for a single forward inference on each city.

\begin{figure}[t]
    \centering
    \includegraphics[width=0.75\linewidth]{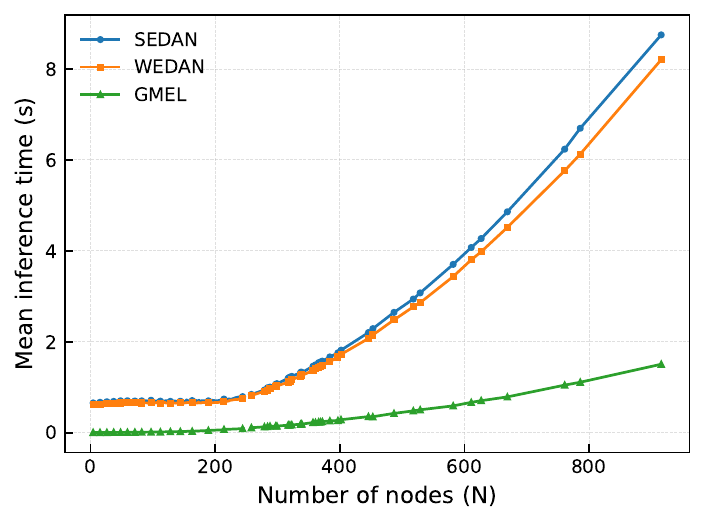}
    \caption{The inference time under different city nodes.}
    \label{fig:inference_time}
\end{figure}

As shown in~\cref{fig:inference_time}, the inference time of all three models increases as the city scale grows. Among them, GMEL exhibits the slowest growth in inference time and consistently maintains the lowest computational cost. In contrast, SEDAN and WEDAN show similar growth trends, and their inference time is noticeably higher than that of GMEL.

For small-scale cities, the difference in inference time among the three models is relatively minor. However, when the number of nodes becomes larger, the diffusion-based models, namely SEDAN and WEDAN, require substantially more inference time than the regression-based model GMEL. Overall, although diffusion models involve multi-step sampling, DDIM acceleration keeps the inference time of SEDAN within an acceptable range even for large-scale cities. Considering its advantages in generation quality and structural consistency, the additional computational cost remains acceptable for applications such as offline OD generation, urban planning, and transportation analysis.

\section{Hyperparameter Analysis}\label{Hyperparameter_Analysis}

\begin{figure*}[t]
    \centering
    \includegraphics[width=\textwidth]{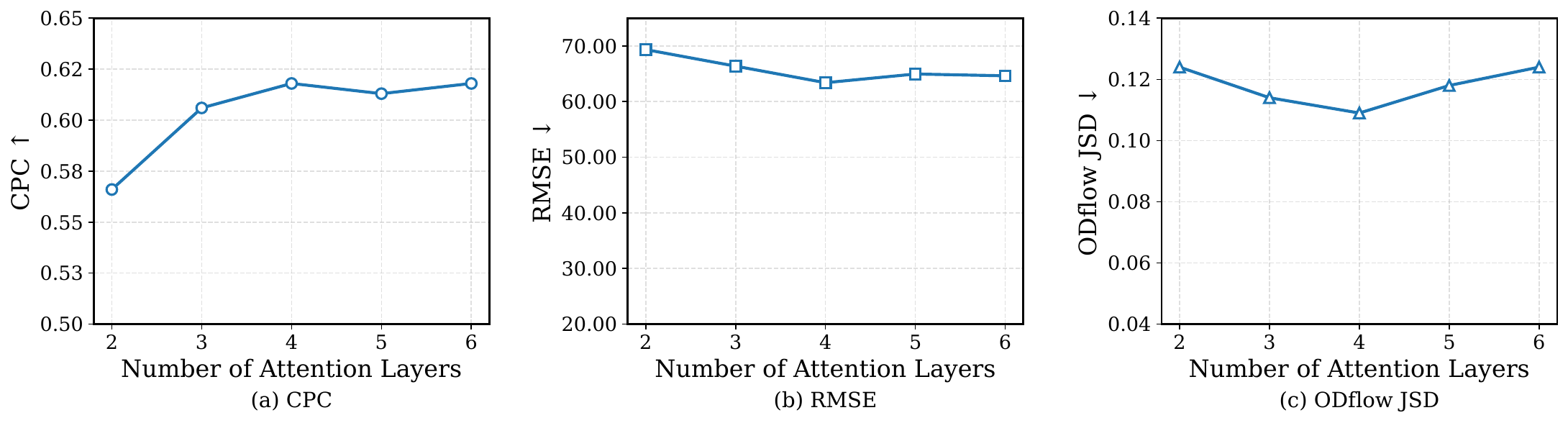}
    \caption{Hyperparameter study on the number of attention layers.}
    \label{fig:hp_layers}
\end{figure*}

\begin{figure*}[t]
    \centering
    \includegraphics[width=\textwidth]{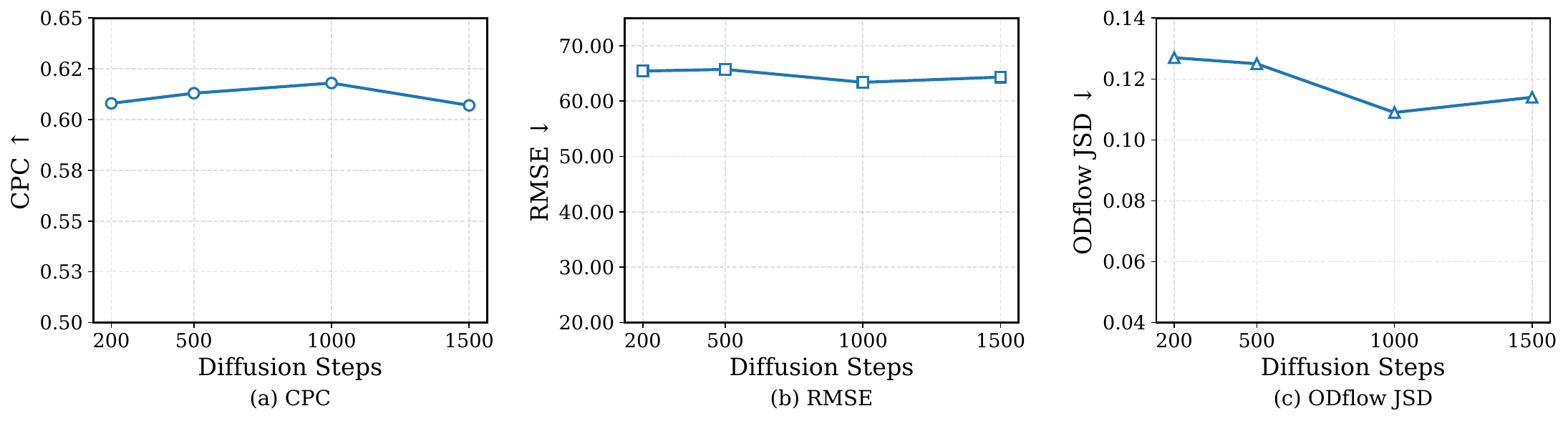}
    \caption{Hyperparameter study on the number of diffusion steps.}
    \label{fig:hp_steps}
\end{figure*}

To examine the effectiveness and robustness of the model configuration, we analyze two key hyperparameters: the number of attention layers in the Graph Transformer and the number of diffusion steps. Model performance is evaluated using CPC, RMSE, and ODFlow JSD, which reflect OD pair consistency, numerical accuracy, and distributional similarity, respectively. The results are shown in Fig.~\ref{fig:hp_layers} and Fig.~\ref{fig:hp_steps}.

For the Graph Transformer, the number of attention layers is varied from 2 to 6. As shown in Fig.~\ref{fig:hp_layers}, increasing the number of layers from 2 to 4 leads to consistent improvements across all metrics, indicating that a moderate model depth is more effective for capturing spatial dependencies. When the number of layers is further increased to 5 or 6, the performance no longer improves and exhibits slight fluctuations. Therefore, four attention layers are adopted as the default setting.

We further study the effect of diffusion steps by setting them to 200, 500, 1000, and 1500. As shown in Fig.~\ref{fig:hp_steps}, the model performs best at 1000 steps, achieving the highest CPC and the lowest RMSE and ODFlow JSD. Further increasing the number of diffusion steps to 1500 yields no additional gain and instead introduces slight fluctuations. Considering the additional computational overhead of longer sampling, 1000 diffusion steps are selected as the default configuration.

\section{SHAP-based Marginal Contribution Analysis}\label{sec:shap}

This appendix describes how we quantify the marginal contribution of regional attributes to the generated OD flows.

\subsection{Explanation Target and Background Construction}
Let $\boldsymbol{X}_i \in \mathbb{R}^M$ denote the attribute vector (demographics and POIs) of a target region $r_i$. The model output $f(\boldsymbol{X}_i)$ represents the predicted average outflow under the fixed urban context.

SHAP quantifies feature importance by measuring how the prediction changes when a feature is treated as missing. Directly setting missing features to zero leads to unrealistic, out-of-distribution inputs. To address this, we adopt a background sampling strategy. Specifically, we construct a global background dataset $\mathcal{B}$ by sampling node attributes from multiple cities. When a feature is withheld, it is replaced with a value drawn from $\mathcal{B}$, ensuring that perturbed inputs remain within the empirical data distribution.

\subsection{KernelSHAP Approximation}
Exact computation of Shapley values requires evaluating all $2^M$ feature subsets, which is infeasible. We therefore employ KernelSHAP to approximate the solution.

Let $\boldsymbol{z}' \in \{0,1\}^M$ be a binary mask indicating which features are retained. For $z'_j = 1$, feature $j$ keeps its original value; otherwise, it is replaced by a background sample. The corresponding expected model output is approximated as:
$$f_x(\boldsymbol{z}') \approx \mathbb{E}_{\boldsymbol{z} \sim \mathcal{B}} \left[ f\left(\boldsymbol{X}_i \odot \boldsymbol{z}' + \boldsymbol{z} \odot (1 - \boldsymbol{z}')\right) \right],$$
where $\odot$ denotes element-wise multiplication.

KernelSHAP fits the following additive surrogate model:
$$g(\boldsymbol{z}') = \phi_0 + \sum_{j=1}^{M} \phi_j z'_j,$$
where $\phi_0$ is the baseline output and $\phi_j$ denotes the SHAP value of feature $j$.

The parameters are estimated by minimizing the weighted squared loss:
$$\mathcal{L} = \sum_{\boldsymbol{z}' \in Z} \pi_x(\boldsymbol{z}') \left( f_x(\boldsymbol{z}') - g(\boldsymbol{z}') \right)^2,$$
where $Z$ is the set of sampled masks, and the Shapley kernel weight is defined as:
$$\pi_x(\boldsymbol{z}') = \frac{M - 1}{\binom{M}{|\boldsymbol{z}'|} |\boldsymbol{z}'| (M - |\boldsymbol{z}'|)}.$$

In practice, KernelSHAP samples feature masks, evaluates the model on background-replaced inputs, and solves the resulting weighted regression to estimate feature contributions.

\end{document}